\documentclass{article}
\usepackage{iclr2027_conference,times}
\iclrfinalcopy
\usepackage[T1]{fontenc}
\usepackage{amsmath,amssymb,graphicx,booktabs,tabularx,placeins,microtype,needspace}
\usepackage[hidelinks]{hyperref}
\usepackage{xurl,flafter,longtable}
\usepackage{colortbl}
\definecolor{interventionShade}{HTML}{688D93}
\title{Early Learning Shapes Later Directions Of Representation Change In Continual Learning}
\author{Yuantao Deng$^{1*}$ \quad Jinnuo Liu$^{1*}$ \quad Kaizhen Tan$^{1}$ \quad Yuechen Liu$^{2}$ \\ \\
\textnormal{$^1$New York University \quad $^2$Xi'an Jiaotong University} \\
\textnormal{$^*$Equal contribution}
}
\newcommand{\R}{\mathbb{R}}
\newcommand{\Usc}{U_{\mathrm{sc}}}
\newcommand{\Ssc}{\mathcal S_{\mathrm{sc}}}
\newcommand{\covg}{\mathcal C}
\begin{document}
\raggedbottom
\maketitle

\begin{abstract}

Representations continually change as a network learns new tasks. We ask whether early representational changes naturally form a geometric structure that continues to shape later learning. We identify a low-dimensional subspace of early representation drift, which we call a \textbf{scaffold}, and test whether it is reused across subsequent tasks. Across four pretrained visual encoders and two datasets, later representational changes consistently favor this early-defined subspace over matched random alternatives. This reuse is history-dependent: when networks experience different early tasks but identical later training inputs, each network preferentially reuses the scaffold induced by its own learning history. The same preference appears in individual optimizer updates, even though the network's dominant local response directions shift away from the original scaffold. Finally, constraining motion within the scaffold slows new-task acquisition more than matched random constraints, while effects on old-task retention are less consistent. In summary, these results suggest that early experience leaves a persistent geometric imprint on how neural networks adapt to future tasks. 
Code is available at \url{https://github.com/YuantaoDeng/latent-scaffold}.
\end{abstract}

\label{sec:abstract}

\section{Introduction}\label{sec:intro}

Continual learning requires neural networks to continually reshape their representations as new tasks arrive. Most work on stability and plasticity asks how previously acquired knowledge
can be preserved despite these changes \citep{Delange_2021, wang2024comprehensivesurveycontinuallearning, Hadsell2020}. Here we ask a different question: \textit{is there stability in the directions of change themselves?} Specifically, we test whether early representational changes reveal a low-dimensional subspace that is repeatedly reused during later learning, and whether restricting motion within this subspace impairs the learning of new tasks.

This question is motivated by evidence that learning is shaped by existing neural geometry. In biological systems, new activity patterns are easier to learn when they lie within previously accessible manifolds, and learned activity can be reassigned to support new behavior \citep{sadtler2014,golub2018}. More broadly, work on representational drift shows that neural representations can change substantially while behavior remains stable \citep{rule2019,driscoll2022,aitken2022}. Recent hippocampal results further link such drift to learning and memory \citep{tang2025}. Overall, these findings suggest that the geometry of representational change, not only the representations themselves, may contain reusable structure. 

\looseness=-1 We study this question by tracking how the representations of fixed anchor inputs move across successive tasks. Whereas singular vector canonical correlation analysis (SVCCA) and centered kernel alignment (CKA) compare representation states \citep{raghu2017,kornblith2019}, we focus on the displacements between states. Across networks that share a pretrained encoder but follow different training histories, we extract the dominant directions of early representational drift and pool them into a fixed low-dimensional subspace, which we call a \emph{scaffold}. We use this term since these directions are established early yet remain available to support later representational change. We then ask whether later learning continues to place disproportionate motion within this scaffold.

Repeated reuse alone, however, does not establish that the scaffold matters for learning. Continual-learning methods often constrain parameters, gradients, or activation subspaces to preserve prior knowledge \citep{kirkpatrick2017,lopezpaz2017,farajtabar2020,saha2021}. We invert this perspective: starting from ordinary replay-based learning \citep{chaudhry2019}, we first identify the directions that learning naturally reuses, and then directly restrict motion along them. This intervention lets us test whether the observed geometry is merely descriptive or functionally involved in acquiring new tasks. 

Across four pretrained visual encoders and two datasets, we find that \textbf{early representational drift defines a subspace that remains preferentially occupied by later changes, even in held-out training histories}. This reuse is history-dependent: when networks share the same initialization and later inputs but experience different early tasks, each network's subsequent drift preferentially aligns with the scaffold induced by its early experience. The preference persists at the level of individual optimizer updates, even as responses to random parameter perturbations become less aligned with the original scaffold. Finally, restricting motion within the scaffold slows new-task acquisition more than energy-matched random constraints, while effects on old-task retention are less systematic.

Our main contributions are:
\begin{itemize}
    \item We introduce a geometric view of continual learning that focuses on the \emph{directions of representational change}, and define a low-dimensional early-learning scaffold from pooled representation drift.
    
    \item We show that this scaffold is persistently reused during later learning and that its reuse depends much on early experience.
    
    \item We connect this persistent geometry to optimization and learning behavior: actual updates continue to favor the scaffold, and constraining motion within it selectively impairs new-task acquisition.
\end{itemize}

\begin{figure}
    \centering
    \includegraphics[width=1\linewidth]{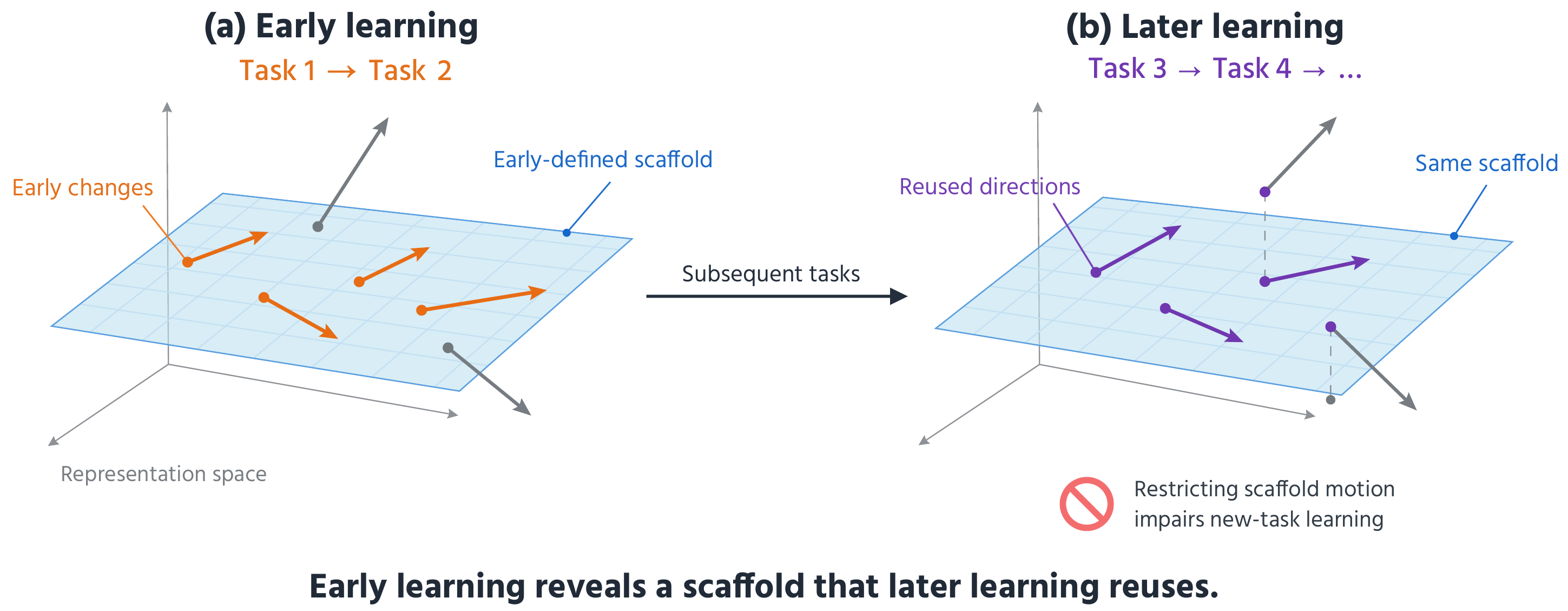}
    \caption{Early learning reveals a reusable scaffold. (a) Early representational changes define a low-dimensional subspace, the scaffold. (b) Later learning preferentially reuses these directions. Restricting motion within the scaffold impairs new-task acquisition more than energy-matched random constraints.}
    \label{fig:overview}
\end{figure}
\section{Related Work}\label{sec:related}

\paragraph{Representational change in continual learning.}
Representation-comparison methods such as SVCCA and CKA quantify how hidden states differ across layers, networks, or training stages \citep{raghu2017,kornblith2019}. Applied to continual learning, such analyses locate forgetting in deeper layers and relate it to task similarity \citep{ramasesh2021}, and show that newly arriving classes cause abrupt representation changes under replay \citep{caccia2022}. Work on representational drift characterizes its geometry in biological and artificial networks \citep{aitken2022} and relates it to ongoing plasticity \citep{vanderveldt2026}. These studies describe how much and where representations change; we ask whether early displacements define a fixed subspace that continues to capture later change.

\paragraph{Early experience and low-dimensional learning dynamics.}
Early training can have lasting effects: temporary input deficits early in training impair final performance despite later training on intact inputs \citep{achille2019}, and parameter gradients concentrate early in a slowly changing subspace spanned by leading Hessian eigenvectors \citep{gurari2018}. In neuroscience, population subspaces formed by earlier learning are reused to accelerate new learning in recurrent networks and primates \citep{goudar2023,tian2026}. Our scaffold is defined in feature space instead of parameter space, and our early-task branches test if its orientation depends on the content of early experience.

\paragraph{Constraining directions of change.}
Many continual-learning methods protect earlier knowledge by restricting update directions, projecting gradients away from directions that affect earlier outputs or activations \citep{farajtabar2020,saha2021}. Scaled Gradient Projection relaxes strict orthogonality because it hinders new learning \citep{saha2023}. Closest to our intervention, \citet{si2026} show that anchoring an old representation during replay suppresses drift and impairs later acquisition. We find that this cost is concentrated in particular directions: restricting motion within the scaffold impairs acquisition more than restricting random directions. 
\section{Defining the Early-Learning Scaffold}\label{sec:scaffold}

We define the \emph{scaffold} as a low-dimensional subspace of early representational drift, then test whether it captures later change in separate training histories.

\subsection{Training Histories and Representational Drift}

We study ResNet-18, DeiT-Tiny, MobileNetV3-Small, and MLP-Mixer-B/16 on CIFAR-10 and CIFAR-100 \citep{he2016,touvron2021,howard2019,tolstikhin2021,krizhevsky2009}. Within each model--dataset condition, networks start from the same pretrained encoder weights, providing common feature coordinates, but follow different task orders and training randomness. We refer to each resulting trajectory as a \emph{training history}. Baseline sequences comprise five two-class tasks on CIFAR-10 and four ten-class tasks on CIFAR-100. Training combines current-task examples with balanced replay of earlier tasks and classifies among all classes seen so far. Details of training are in Appendix~\ref{app:training}.

We track representations of fixed first-task images, called \emph{anchors}. Let $Z_t\in\R^{n\times d}$ contain the $d$-dimensional features of $n$ anchors after task $t$. The centered representational drift between task endpoints $a$ and $b$ is
\begin{equation}
D_{a\to b}=H(Z_b-Z_a),\qquad
H=I_n-\frac1n\mathbf1_n\mathbf1_n^\top.
\label{eq:drift}
\end{equation}
Here $H$ subtracts the mean displacement across anchors from each anchor's displacement. Centering thus removes the translation shared by all anchors and retains changes in their relative representations. Scaffold fitting uses first-task training images; evaluation uses held-out first-task images excluded from both training and replay.

\subsection{Scaffold Construction and Evaluation}

For each condition, we construct the scaffold from $S=6$ estimation histories spanning three task orders with two runs each. Let $V_s\in\R^{d\times r}$ contain the leading $r$ right singular vectors of $D^{(s)}_{1\to2}$, capturing the dominant directions of change during the second task. We average their projection matrices and extract the leading eigenspace:
\begin{equation}
M=\frac1S\sum_{s=1}^S V_sV_s^\top,\qquad
\Usc=\operatorname{eig}_r(M).
\label{eq:scaffold}
\end{equation}
Here $\operatorname{eig}_r$ returns the eigenvectors associated with the largest $r$ eigenvalues. The scaffold is $\Ssc=\operatorname{span}(\Usc)$. This pooling favors directions shared across histories while weighting each history equally, regardless of its drift magnitude.

We choose $r$ based on how well the scaffold captures later drift in estimation histories, relative to a subspace defined by principal components of task-two representations. The selected dimensions range from 8 to 24; Appendix~\ref{app:rank} details the selection procedure and sensitivity analysis. We then fit the scaffold on all six estimation histories and keep its basis and dimension fixed throughout evaluation on separate test histories.

For a nonzero drift matrix $D$ and an orthonormal subspace basis $U$, we define \emph{coverage} as
\begin{equation}
\covg(D,U)=\frac{\|DU\|_F^2}{\|D\|_F^2}.
\label{eq:coverage}
\end{equation}
Coverage is the fraction of centered drift energy captured by the subspace, where energy denotes the squared Frobenius norm. We use \emph{reuse} to mean that later drift occupies the fixed scaffold, regardless of its orientation within that subspace.

To assess whether later drift preferentially occupies the scaffold, we compare its coverage with \emph{classifier-matched random subspaces}. These controls match its dimension and the full spectrum of overlap with the feature span that affects relative class scores, while randomizing orientation within that span and its orthogonal complement. Coverage above this reference therefore indicates a preference beyond that accounted for by dimension and classifier alignment. We average coverage over 128 controls per test history (Appendix~\ref{app:controls}).

\section{A Persistent Scaffold Shaped by Early Experience}\label{sec:recurrence}
Early drift identifies a fixed subspace that captures later change across separate training histories. We first test this persistence across task transitions, then vary early task content to examine how experience shapes the directions that later drift favors.

\subsection{Early Change Predicts Later Representational Drift}
\begin{figure}[!ht]
\centering
\includegraphics[width=\linewidth]{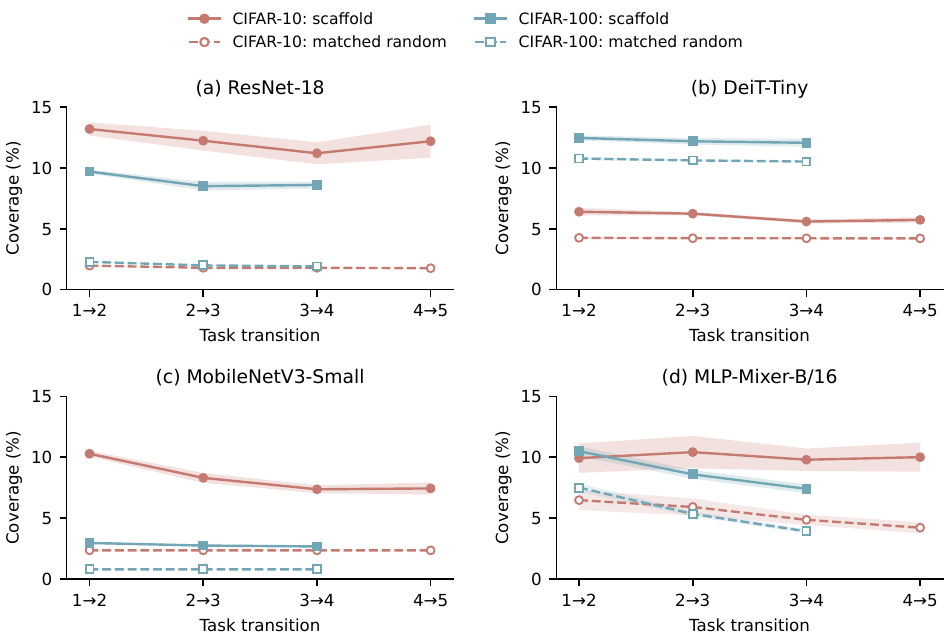}
\caption{\textbf{The same early scaffold remains present in later task drift.} Each panel shows one encoder's absolute drift coverage on CIFAR-10 (muted-coral circles) and CIFAR-100 (grey-blue squares). Solid curves with filled markers show the fixed scaffold; dashed curves with open markers show classifier-matched random controls. The $1\to2$ points test transfer across histories; later points test persistence after further learning. Shaded bands show the standard error for each coverage curve.}
\label{fig:recurrence}
\end{figure}

\textbf{The fixed early scaffold captures more drift than classifier-matched random subspaces at every measured transition in separate test histories} in Figure~\ref{fig:recurrence}, with positive nominal 95\% intervals for all paired coverage differences across the eight model–dataset conditions. The $1\to2$ comparisons test transfer across histories; later transitions test persistence with the same basis and rank.

These histories share an architecture and pretrained encoder. To examine how early experience shapes the recurring geometry, we next vary the second task while holding later training inputs fixed.

\subsection{Controlled Early-Task Branching}\label{sec:content}
We construct paired histories that share the same network state after task $A$. The branches learn alternative second tasks, $B_X$ and $B_Y$, with disjoint class sets, then resume common training. We evaluate drift over the first common task on CIFAR-10 and CIFAR-100; a second common task on CIFAR-10 tests persistence after further shared training. Each model--dataset condition uses one fixed class partition (Appendix~\ref{app:content}).

During common training, each test pair receives the same current and replay examples, with matched augmentations and batch sequences. Replay excludes examples from $B_X$ and $B_Y$. Both branches use the same output class set, and AdamW is reset at task boundaries. 

We fit scaffold bases $U_X$ and $U_Y$ from the respective $1\to2$ drift of six pairs of estimation runs, using the same baseline-selected rank for both branches. We evaluate each branch of six separate test pairs against \emph{both} fixed scaffolds. For later drift $D_X$ and $D_Y$, the own-history advantages are
\begin{equation}
\begin{aligned}
\Delta_X&=\covg(D_X,U_X)-\covg(D_X,U_Y),\\
\Delta_Y&=\covg(D_Y,U_Y)-\covg(D_Y,U_X).
\end{aligned}
\label{eq:content}
\end{equation}
Requiring both contrasts to be positive rules out the possibility that one scaffold simply captures more later drift in both branches. Positive $\Delta_X$ and $\Delta_Y$ together indicate that each branch's later drift favors the scaffold induced by its own early task.

\subsection{Later Drift Favors the Own-History Scaffold}
\begin{figure}[!ht]
\centering
\includegraphics[width=\linewidth]{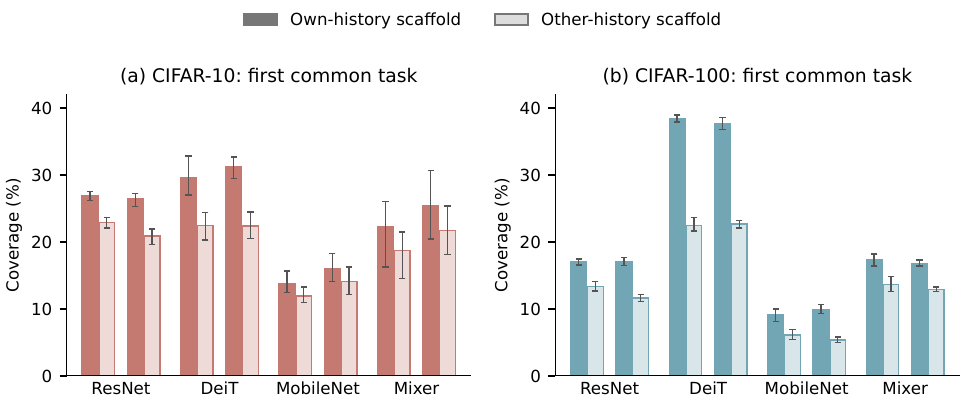}
\caption{\textbf{Later drift favors the own-history scaffold under identical training inputs.} Absolute drift coverage on the first common task (task three) for CIFAR-10 (left) and CIFAR-100 (right). For each model, the left and right bar pairs show test branches $X$ and $Y$, respectively; solid and pale bars use the own-history and other-history scaffolds. Bars average six paired test seeds; error bars show nominal 95\% paired-seed bootstrap intervals, conditional on the fitted scaffolds. Appendix~\ref{app:content} reports the paired own-history contrasts.}
\label{fig:content}
\end{figure}

On the first common task, mean drift coverage is higher in each branch's own-history scaffold than in the other-history scaffold, despite identical later training inputs (Figure~\ref{fig:content}). Both own-history contrasts have nominal 95\% paired-seed intervals entirely above zero in all eight model--dataset conditions. This mean preference persists through CIFAR-10's second common task in all four models (Figure~\ref{fig:content_second} in Appendix~\ref{app:content}).

History-specific preference coexists with partial alignment between the scaffolds. For orthonormal rank-$r$ bases $U$ and $V$, we measure orientation overlap as
\begin{equation}
\mathcal O(U,V)=\frac{\|U^\top V\|_F^2}{r}\in[0,1].
\label{eq:overlap}
\end{equation}
Overlap is zero for orthogonal subspaces and one for identical subspaces. Unlike coverage, it weights retained directions equally rather than by observed drift energy. Across models, $\mathcal O(U_X,U_Y)$ ranges from 44.2--58.7\% on CIFAR-10 and 14.2--58.9\% on CIFAR-100 (Appendix~\ref{app:content}).

\section{Learning Actively Reuses the Scaffold}\label{sec:updates}
The fixed scaffold captures later task-level drift. We now ask whether it also captures the feature motion produced during learning, even as the network's local responses to parameter changes evolve. We compare the scaffold with dominant responses to random parameter perturbations and with motion from actual optimizer updates.

\subsection{Local Response Alignment Declines after Early Learning}

At each task endpoint, we apply small perturbations along sixteen isotropically sampled encoder-parameter directions and measure centered feature responses on fixed anchors. Classifier weights and normalization buffers remain fixed. The leading response directions define a local \emph{response subspace}. We pool response projectors across the six estimation histories at each endpoint, using the scaffold's fixed rank, and measure alignment with the scaffold by the orientation overlap in Equation~\ref{eq:overlap}. Appendix~\ref{app:response} gives the finite-difference definition and local linear interpretation.

\begin{figure}[!ht]
\centering
\includegraphics[width=\linewidth]{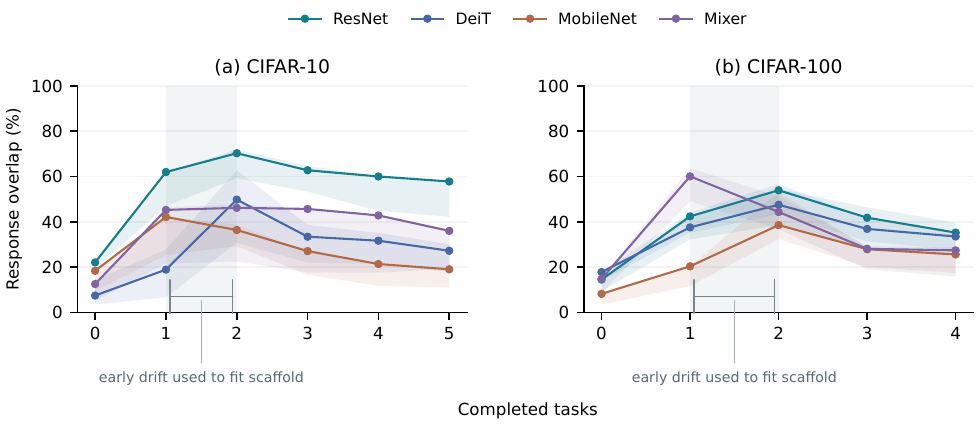}
\caption{\textbf{Response alignment declines after early learning.} Orientation overlap between the fixed scaffold and pooled local response subspaces on CIFAR-10 (a) and CIFAR-100 (b). The pale region and bracket mark the $1\to2$ interval used for scaffold construction. Shaded bands show nominal 95\% intervals from 2,000 estimation-order/run resamples that refit both subspaces.}
\label{fig:response}
\end{figure}

Alignment rises early, with point estimates peaking at task one or two (Figure~\ref{fig:response}). It is lower at the final task than at task two in all eight baseline conditions, with nominal 95\% intervals for these final-minus-task-two contrasts entirely below zero. Task-level drift nevertheless retains a coverage advantage in the scaffold over matched controls.

Isotropic perturbations probe local responses, whereas optimizer updates follow the training loss. We next test whether the feature motion of those updates continues to favor the scaffold.

\subsection{Optimizer Updates Continue to Favor the Scaffold}

We measure centered feature motion from actual AdamW steps near the beginning, middle, and end of the final task. For each of the eight conditions, we resume twelve test histories from pre-final-task states under the same five-epoch replay protocol, with the scaffold and classifier-matched reference fixed before evaluation. Coverage is averaged over three neighboring successful steps within each history and window (Appendix~\ref{app:updates}).

\begin{figure}[!ht]
\centering
\includegraphics[width=\linewidth]{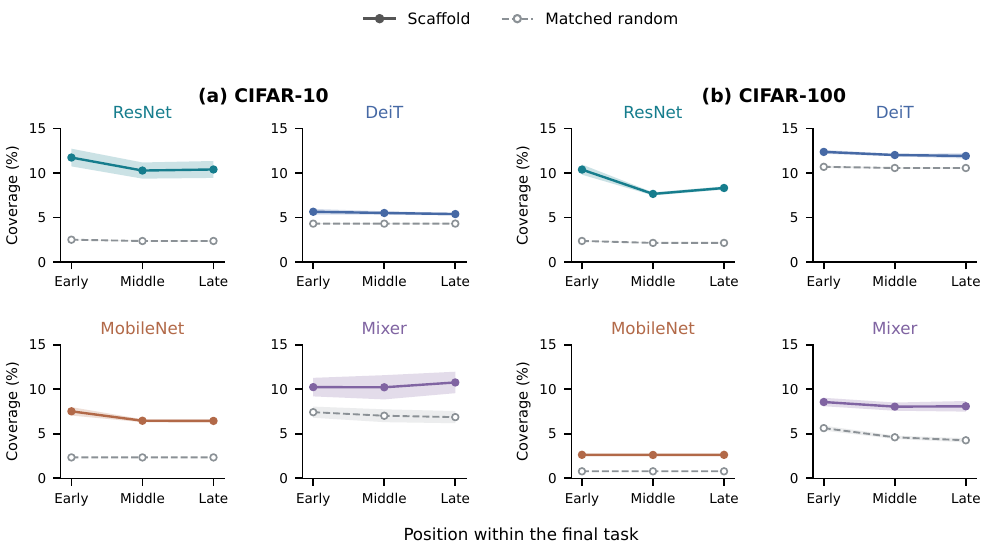}
\caption{\textbf{Sampled optimizer updates continue to favor the scaffold.} Absolute AdamW feature-motion coverage in early, middle, and late windows of the final task on CIFAR-10 (a) and CIFAR-100 (b). Solid curves show scaffold coverage; dashed curves show classifier-matched random coverage. Each history contributes the mean coverage of three neighboring updates, and curves average twelve test histories per condition. Shaded bands show $\pm$ one hierarchical order/run bootstrap standard error, conditional on the fitted scaffold. Paired coverage-gain intervals are reported in Appendix~\ref{app:updates}.}
\label{fig:updates}
\end{figure}

Mean update coverage exceeds the matched reference in all three windows of every condition, with nominal 95\% paired hierarchical intervals for these gains entirely above zero (Figure~\ref{fig:updates}). Gains are also positive in 859 of the 864 sampled AdamW steps. A training-gradient control matched to the norm of the AdamW encoder step yields positive gain intervals in every condition and window (Appendix~\ref{app:updates}).

A complementary analysis uses scaffold-projected drift over the baseline final task to reconstruct class-score and loss changes under a frozen classifier. Gains over matched controls vary across conditions (Appendix~\ref{app:readout}).

Together, these measurements show that \textbf{sampled optimizer updates continue to favor the scaffold as dominant local responses become less aligned with it}. We next restrict motion within the baseline scaffold and measure effects on new-task acquisition and old-task retention.

\section{Restricting the Scaffold Impairs New-Task Acquisition}\label{sec:intervention}
We compare learning when a quadratic penalty restricts feature motion within the pooled baseline scaffold or random subspaces. The comparison tests how constraining this space affects new-task acquisition. We run paired continuations of the final task for all four encoders on both datasets, using three predefined task orders and two runs per order. Within each history, normal training and all constrained branches share their initial state and current/replay input stream. For old-memory features $Z_\theta^{\rm mem}$, pre-task references $Z_0^{\rm mem}$, and minibatch-centering matrix $H_m$, we train with
\begin{equation}
\mathcal L(\theta)=\mathcal L_{\rm current}+\mathcal L_{\rm replay}
+\frac{\lambda}{mr}\big\|H_m(Z_\theta^{\rm mem}-Z_0^{\rm mem})U\big\|_F^2.
\label{eq:suppress}
\end{equation}
\begin{samepage}
The first two terms are mean classification losses on current and replay examples. In the penalty, $m$ is the replay minibatch size (32) and $H_m$ subtracts the mean feature displacement. The columns of $U$ span the fixed scaffold or a random subspace. All bases have the same rank $r$. Dividing by $mr$ averages squared projected motion over examples and retained directions; $\lambda$ sets the penalty coefficient. We average the two random-constraint branches within each history. The classifier remains trainable, and held-out evaluation images never enter the penalty. Acquisition is new-task accuracy averaged over epochs one through five.
\par
\end{samepage}

We use two random comparisons. The first applies $\lambda=100$ to both the scaffold and uniform random subspaces. The second matches the drift energy removed along the constrained directions. For this comparison, we first measure drift on old training-memory examples after normal final-task training. We construct two random subspaces that capture the same amount of this drift energy as the scaffold. We then keep the scaffold coefficient at 100 and adjust each random coefficient until it removes a similar amount of energy along its own directions. We call this the \emph{energy-matched} comparison. Appendix~\ref{app:energy_matched} gives the construction and matching results.

\begin{table}[!ht]
\centering
\caption{\textbf{Acquisition costs persist after matching removed drift energy.} New-task acquisition averages accuracy over epochs one through five; old-task retention averages final accuracy across all old tasks. Entries show mean accuracy (\%) and nominal 95\% hierarchical order/run bootstrap intervals (20,000 draws). Each condition averages six paired histories and averages random branches within each history. Scaffold and same-$\lambda$ random use $\lambda=100$; energy-matched random matches removed drift energy. Within each row and metric, darker shading indicates higher accuracy among the three methods. Shading shows relative ranks, not statistical significance. Appendix~\ref{app:intervention} reports paired scaffold-minus-random contrasts.}
\label{tab:intervention}
\begingroup
\setlength{\tabcolsep}{3.4pt}
\renewcommand{\arraystretch}{1.25}
\footnotesize
\begin{tabular}{llcccccc}
\toprule
 & & \multicolumn{3}{c}{New-task acquisition (\%)} & \multicolumn{3}{c}{Old-task retention (\%)}\\
\cmidrule(lr){3-5}\cmidrule(lr){6-8}
Model & Data & Scaffold & Random & Random & Scaffold & Random & Random\\
 & & & Same $\lambda$ & Energy matched & & Same $\lambda$ & Energy matched\\
\midrule
ResNet & C10 & \cellcolor{interventionShade!10}\shortstack{67.43\\{\scriptsize [61.60, 75.33]}} & \cellcolor{interventionShade!24}\shortstack{74.40\\{\scriptsize [69.00, 82.17]}} & \cellcolor{interventionShade!40}\shortstack{89.02\\{\scriptsize [85.85, 93.85]}} & \cellcolor{interventionShade!10}\shortstack{68.83\\{\scriptsize [66.42, 70.75]}} & \cellcolor{interventionShade!24}\shortstack{69.23\\{\scriptsize [67.67, 70.83]}} & \cellcolor{interventionShade!40}\shortstack{71.29\\{\scriptsize [69.94, 72.50]}} \\
 & C100 & \cellcolor{interventionShade!10}\shortstack{77.44\\{\scriptsize [76.16, 78.88]}} & \cellcolor{interventionShade!24}\shortstack{83.51\\{\scriptsize [82.38, 84.60]}} & \cellcolor{interventionShade!40}\shortstack{91.34\\{\scriptsize [90.15, 92.44]}} & \cellcolor{interventionShade!40}\shortstack{61.41\\{\scriptsize [59.03, 64.84]}} & \cellcolor{interventionShade!24}\shortstack{61.10\\{\scriptsize [58.78, 64.04]}} & \cellcolor{interventionShade!10}\shortstack{59.62\\{\scriptsize [57.67, 61.68]}} \\
\midrule
DeiT & C10 & \cellcolor{interventionShade!10}\shortstack{70.60\\{\scriptsize [64.57, 79.40]}} & \cellcolor{interventionShade!24}\shortstack{70.95\\{\scriptsize [64.70, 79.80]}} & \cellcolor{interventionShade!40}\shortstack{82.32\\{\scriptsize [76.00, 87.88]}} & \cellcolor{interventionShade!10}\shortstack{68.96\\{\scriptsize [65.25, 71.88]}} & \cellcolor{interventionShade!24}\shortstack{69.31\\{\scriptsize [66.25, 71.62]}} & \cellcolor{interventionShade!40}\shortstack{71.79\\{\scriptsize [68.87, 73.67]}} \\
 & C100 & \cellcolor{interventionShade!10}\shortstack{82.81\\{\scriptsize [81.24, 84.19]}} & \cellcolor{interventionShade!24}\shortstack{83.60\\{\scriptsize [82.14, 84.69]}} & \cellcolor{interventionShade!40}\shortstack{91.18\\{\scriptsize [89.36, 92.48]}} & \cellcolor{interventionShade!40}\shortstack{56.83\\{\scriptsize [54.00, 59.93]}} & \cellcolor{interventionShade!24}\shortstack{56.53\\{\scriptsize [53.17, 60.41]}} & \cellcolor{interventionShade!10}\shortstack{56.32\\{\scriptsize [54.58, 58.31]}} \\
\midrule
MobileNet & C10 & \cellcolor{interventionShade!10}\shortstack{76.13\\{\scriptsize [58.03, 84.73]}} & \cellcolor{interventionShade!24}\shortstack{80.88\\{\scriptsize [69.95, 86.85]}} & \cellcolor{interventionShade!40}\shortstack{82.60\\{\scriptsize [72.53, 87.58]}} & \cellcolor{interventionShade!24}\shortstack{58.33\\{\scriptsize [51.38, 63.92]}} & \cellcolor{interventionShade!40}\shortstack{59.04\\{\scriptsize [54.37, 62.75]}} & \cellcolor{interventionShade!10}\shortstack{57.67\\{\scriptsize [48.33, 63.00]}} \\
 & C100 & \cellcolor{interventionShade!10}\shortstack{74.17\\{\scriptsize [69.74, 76.99]}} & \cellcolor{interventionShade!24}\shortstack{75.04\\{\scriptsize [68.93, 78.71]}} & \cellcolor{interventionShade!40}\shortstack{80.93\\{\scriptsize [73.13, 85.49]}} & \cellcolor{interventionShade!40}\shortstack{50.34\\{\scriptsize [47.20, 54.38]}} & \cellcolor{interventionShade!24}\shortstack{50.16\\{\scriptsize [45.20, 54.99]}} & \cellcolor{interventionShade!10}\shortstack{49.71\\{\scriptsize [46.14, 54.03]}} \\
\midrule
Mixer & C10 & \cellcolor{interventionShade!10}\shortstack{64.17\\{\scriptsize [58.87, 71.00]}} & \cellcolor{interventionShade!24}\shortstack{72.48\\{\scriptsize [64.33, 83.85]}} & \cellcolor{interventionShade!40}\shortstack{72.50\\{\scriptsize [61.90, 85.90]}} & \cellcolor{interventionShade!24}\shortstack{57.58\\{\scriptsize [53.67, 61.21]}} & \cellcolor{interventionShade!10}\shortstack{55.79\\{\scriptsize [51.48, 61.13]}} & \cellcolor{interventionShade!40}\shortstack{59.33\\{\scriptsize [55.08, 63.62]}} \\
 & C100 & \cellcolor{interventionShade!10}\shortstack{81.01\\{\scriptsize [80.15, 81.72]}} & \cellcolor{interventionShade!24}\shortstack{84.50\\{\scriptsize [83.09, 85.55]}} & \cellcolor{interventionShade!40}\shortstack{88.18\\{\scriptsize [87.08, 89.20]}} & \cellcolor{interventionShade!40}\shortstack{48.53\\{\scriptsize [45.93, 51.58]}} & \cellcolor{interventionShade!10}\shortstack{44.64\\{\scriptsize [40.93, 47.56]}} & \cellcolor{interventionShade!24}\shortstack{46.64\\{\scriptsize [44.03, 49.59]}} \\
\bottomrule

\end{tabular}
\par\vspace{3pt}
{\scriptsize C10: CIFAR-10; C100: CIFAR-100. Shading: \colorbox{interventionShade!10}{low}\enspace\colorbox{interventionShade!24}{medium}\enspace\colorbox{interventionShade!40}{high}.}
\endgroup
\end{table}

\begin{samepage}
With the same coefficient, scaffold suppression lowers mean acquisition accuracy in all eight conditions (Table~\ref{tab:intervention}). Five paired 95\% intervals lie below zero; three span zero. The reductions range from 0.35 to 8.32 percentage points. This comparison shows an acquisition cost with less consistent evidence across conditions than the energy-matched comparison.

With energy matching, acquisition is lower under scaffold suppression in all eight conditions, with all eight intervals below zero. The reductions range from 6.47 to 21.58 percentage points. Averaging accuracy across training epochs captures learning differences that final accuracy alone could obscure. The two constraints remove similar amounts of projected drift energy from old training-memory examples. Acquisition costs differ despite the matched reduction in projected drift energy.
\par
\end{samepage}

Old-task retention shows no corresponding pattern of consistent loss. With the same coefficient, no interval for final accuracy across all old tasks lies below zero, and one is positive. With energy matching, retention improves for ResNet and Mixer on CIFAR-100 and declines for DeiT on CIFAR-10. The remaining intervals span zero. The observed cost is therefore more consistent for new-task acquisition than for old-task retention. Acquisition averages accuracy during training, whereas retention measures final performance on earlier tasks. The intervention thus produces a consistent acquisition cost alongside varied retention outcomes.

Appendix~\ref{app:intervention} reports both random comparisons and the fixed-coefficient results at $\lambda=10$ and 100. Appendix~\ref{app:accuracy} provides absolute task accuracies and intervention epoch tables. Experiments without replay appear in Appendix~\ref{app:noreplay}.

\section{Discussion}

\textbf{Plasticity has a history-dependent geometry.}
Continual learning is usually framed as a trade-off between preserving what
representations encode and allowing them to change. Our results add a third
element: \textit{the directions of change are themselves inherited from learning
history}. What persists is a route for change instead of a representation,
since representations can change substantially while continuing to reuse the
same directions.

\textbf{Implications for continual-learning methods.}
Methods that restrict gradient or feature directions
\citep{farajtabar2020,saha2021} are usually motivated by retention. Our
intervention suggests that directions which carried past change may also be
those that future learning relies on. Protection could therefore be more
selective, separating directions that encode old tasks from directions that
carry change, and scaffold coverage offers a direct measure of the latter. It
could also serve as a diagnostic in studies of plasticity loss
\citep{lyle2023,dohare2024}, testing whether declining plasticity coincides
with reduced use of historically preferred directions.

\textbf{Early experience and curricula.}
If early tasks set the directions that later learning favors, the content and
order of early tasks in a curriculum, or early fine-tuning stages after
pretraining, may shape adaptation long afterward. This extends the lasting
effects of early training \citep{achille2019} from final performance to the
geometry of later change.

\textbf{Neural representational drift.}
Drift in neural circuits is often treated as variability that behavior must
tolerate. Our results suggest that drift may instead be structured by history,
concentrating in subspaces set by earlier learning. Unlike the intrinsic
manifolds that constrain learning \citep{sadtler2014}, such a scaffold is
defined by displacements rather than by activity covariance. Together with
evidence for schema reuse \citep{goudar2023,tian2026} and the learning cost of
stabilizing hippocampal place fields \citep{tang2025}, this raises a testable
question: does later drift in recorded populations preferentially occupy a
subspace defined by early learning, and does constraining it impair learning?

\textbf{Scope and Limitations.}
Our experiments use pretrained visual classifiers in shared feature
coordinates, with one class partition per model and dataset in the content
study, and intervention effects vary across conditions. Training from scratch,
other domains, and other learning regimes would test the generality of these
findings.
\section{Conclusion}
In the tested visual learners, early representational change reveals a fixed low-dimensional scaffold. Later drift and optimizer updates continue to favor this subspace. Controlled early tasks affect which scaffold later drift favors. Constraining the pooled baseline scaffold impairs new-task acquisition relative to energy-matched random constraints. Old-task retention varies across conditions. These findings characterize persistent subspaces in representation change and the learning costs of restricting motion within them.

\bibliography{references}
\bibliographystyle{iclr2027_conference}
\clearpage
\appendix
\section{Training and Evaluation Design}\label{app:training}
\subsection{Shared-coordinate baseline}
The four encoder feature dimensions are 512 (ResNet-18), 192 (DeiT-Tiny), 1024 (MobileNetV3-Small), and 768 (MLP-Mixer-B/16). Each model--dataset condition has a separately fitted scaffold. The pretrained configurations are ResNet-18, DeiT-Tiny, MobileNetV3-Small with ImageNet-1k weights, and Mixer-B/16 with ImageNet-21k pretraining followed by ImageNet-1k training. Shared encoder weights provide common feature coordinates within a condition; classifiers and training randomness vary across baseline histories. Cross-model comparisons describe these pretrained configurations as a whole.

Baseline sequences contain five two-class tasks for CIFAR-10 and four ten-class tasks for CIFAR-100. Each CIFAR-100 baseline trajectory contains forty classes. The first 450 images per selected class are training images, and the last fifty are held out. Fixed first-task training anchors are selected by the balanced-memory rule; held-out first-task anchors never participate in training or replay. This gives 500 fitting anchors and 100 held-out anchors on CIFAR-10, and 500 fitting and 500 held-out anchors on CIFAR-100. Different histories can share training images and classes. The estimation and test histories are independent runs within shared training content.

AdamW uses learning rate $3\times10^{-4}$ and weight decay $0.05$, with five epochs per task and optimizer state reset at task boundaries. The batch size is 64. After the first task, batches contain 32 current and 32 replay examples; the two mean cross-entropies are added. The replay memory holds 500 examples balanced over seen classes, with a fixed within-run random priority determining membership. Predictions compete over all classes seen so far.

Training applies random crops with four-pixel padding and random horizontal flips. Inputs are normalized using ImageNet channel statistics and resized to $224\times224$ by antialiased bilinear interpolation. The same preprocessing is used for all four models. Baseline training and sequential feature extraction use mixed precision; parameter-response and update-field evaluation use FP32 and deterministic preprocessing. Mixer uses activation recomputation to reduce training memory requirements.

\subsection{Sampling and uncertainty}
Each baseline condition uses three estimation task orders with two runs per order, and six separate test orders with two runs per order. CIFAR-10 estimation order identifiers are 2003, 2011, and 2017; test identifiers are 2101, 2111, 2129, 2131, 2137, and 2141. CIFAR-100 estimation identifiers are 1009, 1013, and 1019; test identifiers are 1103, 1109, 1117, 1123, 1129, and 1151. These identify the pseudorandom class-order construction in the experiment sources.

Test means weight task orders equally. A hierarchical bootstrap resamples orders, then runs within each selected order; paired contrasts use the same resamples for both quantities. Baseline drift and readout intervals use 20,000 draws, while multi-position update intervals use 10,000. Geometry intervals use 2,000 estimation-order/run resamples and refit both pooled subspaces jointly. Geometry intervals describe variation in subspace fitting, while test-history intervals describe variation in evaluation histories. Reported intervals are nominal 95\% intervals without multiplicity adjustment.

The early-content experiments instead have one fixed class partition and six independent paired test seeds per model and dataset. Their 10,000 bootstrap draws resample paired seeds, without an order hierarchy. Baseline drift, readout, content, actual-update, and no-replay test intervals condition on the fitted scaffolds. Resampling units are training histories or paired seeds; anchors and neighboring steps are summarized within these units.

\section{Subspace Construction and Random Controls}\label{app:controls}
\subsection{Why projector pooling selects recurring directions}
Write the singular value decomposition of centered early drift as $D_s=A_s\Sigma_s B_s^\top$, where $\Sigma_s$ contains decreasing singular values and $A_s,B_s$ have orthonormal columns. The leading $r$ columns $V_s$ of $B_s$ maximize $\|D_sV_s\|_F^2$ among rank-$r$ orthonormal bases. Their projector $V_sV_s^\top$ is invariant to sign flips and rotations of the basis. With the mean projector $M$ defined in Equation~\ref{eq:scaffold}, an orthonormal basis for the pooled scaffold solves
\begin{equation}
\Usc\in\underset{U^\top U=I_r}{\arg\max}\ \operatorname{tr}(U^\top MU)
=\underset{U^\top U=I_r}{\arg\max}\ \frac1S\sum_{s=1}^S\|V_s^\top U\|_F^2.
\end{equation}
Here $U^\top U=I_r$ requires perpendicular unit columns, $\operatorname{tr}$ sums a matrix's diagonal entries, and $\arg\max$ denotes the bases attaining the largest objective. The right-hand side averages squared dot products between the candidate directions and each history's basis. To see why leading eigenvectors solve this problem, express $U$ in an eigenbasis of $M$: the objective is a weighted sum of the eigenvalues, with each weight between zero and one and weights summing to $r$. It is largest when all weight is placed on the largest $r$ eigenvalues. Thus pooling favors agreement across histories independently of displacement magnitudes. A tie at the boundary can make the maximizing subspace nonunique.

For $P_U=UU^\top$, orthogonality gives
\begin{equation}
\|DP_U\|_F^2=\|DU\|_F^2,\qquad
\|D\|_F^2=\|DP_U\|_F^2+\|D(I_d-P_U)\|_F^2.
\end{equation}
Thus coverage lies in $[0,1]$ when $D\ne0$. Centering decomposes total displacement as $Z_b-Z_a=D+\mathbf1_n\mu^\top$, with $\mu=(Z_b-Z_a)^\top\mathbf1_n/n$. The centered and mean components are orthogonal: the sum of their entrywise products is zero, so their squared energies add.

\subsection{Classifier-matched reference}
Let $W\in\R^{c\times d}$ be the classifier weights for the relevant $c$ classes. Subtract the row mean, and let $Q_W$ be an orthonormal basis for the resulting row space, of dimension $h$. This feature span changes relative class scores. Random rank-$r$ subspaces match the eigenvalues of $\Usc^\top Q_WQ_W^\top\Usc$, preserving the overlap of their directions with this span while randomizing orientation within it and its perpendicular complement. These eigenvalues are squared cosines of the angles between the subspaces: a value of one denotes a shared direction and zero a perpendicular direction. Matching all these values preserves more information than their sum alone.

Baseline temporal controls use the task-2 classifier over seen classes. Baseline readout and multi-position update controls use the old-class classifier immediately before the final task. The baseline averages 128 random draws per test history. In no-replay tests, the reference uses the classifier at the start of the measured transition or task and evaluates the matched distribution's exact expected projection energy.

To see that expectation, write $\tau=\operatorname{tr}(\Usc^\top Q_WQ_W^\top\Usc)$. Write $P_{\mathrm{rand}}$ for a random basis's projection matrix and $\mathbb E$ for its average over random draws. Since no orientation within either span is preferred,
\begin{equation}
\mathbb E[P_{\mathrm{rand}}]=\frac{\tau}{h}Q_WQ_W^\top+
\frac{r-\tau}{d-h}(I_d-Q_WQ_W^\top).
\end{equation}
Expected coverage is $\operatorname{tr}(D^\top D\,\mathbb E[P_{\mathrm{rand}}])/\|D\|_F^2$. Degenerate empty spans are handled by their remaining component. This expectation concerns linear projection energy; nonlinear readout reconstructions use their recorded random controls. Uniform random subspaces without classifier matching instead have expected coverage $r/d$.

Final-transition coverage on CIFAR-100 is 8.60\% for ResNet, 12.07\% for DeiT, 2.67\% for MobileNet, and 7.40\% for Mixer.

\section{Dimension Selection and Sensitivity}\label{app:rank}
The retained dimension $r$ is selected using only the six estimation histories. For each candidate integer rank starting at a specified minimum, leave one of the three task orders out and fit the scaffold from the four runs in the other two orders. The representation-variance reference is constructed in the same way from the leading principal components of task-two feature states. Principal component analysis (PCA) supplies a representation-variance reference for selecting the rank. These principal components describe task-two feature states, after early training. The selection score calibrates the measurement dimension on estimation histories.

For estimation history $s$, let $o(s)$ denote its order. The subscript $-o(s)$ means that both runs of that order were excluded from fitting. With $T$ the final estimation task, define
\begin{align}
g(r)&=\frac1S\sum_{s=1}^S\left[\covg(D^{(s)}_{T-1\to T},U_{{\rm sc},-o(s)}^{(r)})
-\covg(D^{(s)}_{T-1\to T},U_{{\rm var},-o(s)}^{(r)})\right],\notag\\
r_*&=\min\{r\geq r_{\min}:g(r)>0\}.
\label{eq:rank_select}
\end{align}
Here $r_{\min}$ is the minimum admissible rank, set to eight, and $S$ is the number of estimation histories. Candidate ranks are positive integers within the feature dimension. Each history contributes its held-out final-transition coverage difference once. The rule chooses the smallest candidate with positive mean gain, then refits on all six histories. It returns no rank if no candidate qualifies. The selected rank sets the dimension of the measurement subspace. A positive mean difference determines eligibility, independently of statistical significance. Table~\ref{tab:rank_scores} reports the score used to select each dimension.

\begin{table}[htbp]
\centering
\caption{Fixed scaffold dimensions. The same dimensions are retained in all follow-up conditions, including each pair of content branches and separately fitted no-replay scaffolds.}
\begin{tabular}{lrrr}\toprule
Model & Feature dimension & CIFAR-10 rank & CIFAR-100 rank\\\midrule
ResNet-18 & 512 & 8 & 8\\
DeiT-Tiny & 192 & 8 & 20\\
MobileNetV3-S & 1024 & 24 & 8\\
MLP-Mixer-B/16 & 768 & 8 & 8\\\bottomrule
\end{tabular}
\end{table}

\begin{table}[htbp]
\centering\small
\caption{Rank-selection scores on estimation histories. The score $100g(r_*)$ is a coverage difference in percentage points. A positive mean determines eligibility. Each value averages six held-out estimation histories at the selected rank.}
\label{tab:rank_scores}
\begin{tabular}{llrr}\toprule
Model & CIFAR & $r_*$ & Selection difference (pp)\\\midrule
ResNet-18 & 10 & 8 & 2.1744\\
ResNet-18 & 100 & 8 & 1.8322\\
DeiT-Tiny & 10 & 8 & 1.3402\\
DeiT-Tiny & 100 & 20 & 0.0147\\
MobileNetV3-S & 10 & 24 & 0.0362\\
MobileNetV3-S & 100 & 8 & 0.1800\\
MLP-Mixer-B/16 & 10 & 8 & 0.9611\\
MLP-Mixer-B/16 & 100 & 8 & 1.3720\\
\bottomrule

\end{tabular}
\end{table}

The minimum rank sets a compact measurement budget. We repeated the selection with minimum ranks five, six, seven, and eight. The selected scaffolds retain a positive coverage advantage over uniform-random directions in all eight conditions at every tested minimum (Table~\ref{tab:lower_bound}). This qualitative result is stable across the nearby lower-bound choices; the main experiments use the ranks selected with a minimum of eight.

\begin{table}[htbp]
\centering\small
\caption{Sensitivity to the minimum admissible rank. The four middle columns give the selected rank after changing only the minimum. The final column gives the smallest held-out estimation coverage gain over the uniform-random expectation across these four choices (percentage points).}
\label{tab:lower_bound}
\begin{tabular}{llrrrrr}\toprule
 & & \multicolumn{4}{c}{Minimum admissible rank} & Minimum gain\\
Model & CIFAR & 5 & 6 & 7 & 8 & over random\\\midrule
ResNet-18 & 10 & 5 & 6 & 7 & 8 & 11.24\\
ResNet-18 & 100 & 5 & 6 & 7 & 8 & 3.83\\
DeiT-Tiny & 10 & 5 & 6 & 7 & 8 & 7.54\\
DeiT-Tiny & 100 & 20 & 20 & 20 & 20 & 4.16\\
MobileNetV3-S & 10 & 24 & 24 & 24 & 24 & 4.50\\
MobileNetV3-S & 100 & 5 & 6 & 7 & 8 & 1.76\\
MLP-Mixer-B/16 & 10 & 5 & 6 & 7 & 8 & 5.58\\
MLP-Mixer-B/16 & 100 & 5 & 6 & 7 & 8 & 3.78\\
\bottomrule

\end{tabular}
\end{table}
\FloatBarrier

\section{Paired Early-Content Experiments}\label{app:content}
\subsection{Content, training, and matching}
For each model, two training histories share task $A$ and then learn alternative tasks drawn from disjoint class sets. The symbols $B_X$ and $B_Y$ identify complete, class-disjoint tasks occupying the same position in the alternative histories. Within each seed, task $A$ is trained once and its complete state is copied to the two histories. On CIFAR-10, $A$ contains deer and frog, $B_X$ truck and ship, and $B_Y$ horse and airplane. Both branches subsequently learn bird and automobile, then cat and dog. On CIFAR-100, four disjoint ten-class groups define $A$, $B_X$, $B_Y$, and a single common later task $C$. Each dataset uses one fixed class partition per model.

CIFAR-10 uses the class groups $A=[4,6]$, $B_X=[9,8]$, $B_Y=[7,0]$, $C=[2,1]$, and $D=[3,5]$. The branches follow $A\to B_X\to C\to D$ and $A\to B_Y\to C\to D$. Each symbol denotes a complete class-defined task. The alternative histories use complete tasks with disjoint class sets in the second position. On CIFAR-100, the four fixed ten-class groups are listed below.
\begin{center}
\begin{tabular}{ll}\toprule
Task & CIFAR-100 class identifiers\\\midrule
$A$ & 77, 55, 7, 78, 87, 38, 94, 24, 92, 23\\
$B_X$ & 12, 44, 42, 71, 20, 22, 99, 81, 75, 32\\
$B_Y$ & 37, 28, 48, 69, 10, 74, 51, 8, 47, 64\\
$C$ & 34, 39, 62, 72, 65, 93, 30, 95, 90, 19\\\bottomrule
\end{tabular}
\end{center}
These groups use blocks of baseline order 1103, fixed before the new measurements. Each CIFAR-100 branch contains thirty training classes. During common task $C$, both classifiers compete over the same forty-class union.

Within each paired seed, the full state after $A$ is copied to both branches. Each task retains five epochs, 450 images per class, batch size 64, learning rate $3\times10^{-4}$, weight decay $0.05$, and task-boundary AdamW reset. Common task $C$ replays only $A$; CIFAR-10 task $D$ replays $A$ and $C$. Distinctive $B_X$ and $B_Y$ examples are absent from later replay. Saved common-input hashes verify matching augmented tensors, image IDs, and replay IDs. Each branch computes gradients at the state inherited from its own early training.

Each condition uses six paired estimation runs and six separate paired test runs. On CIFAR-10, 500 first-task training anchors fit the scaffolds and 100 held-out anchors measure drift; CIFAR-100 uses 500 of each. Estimation and test runs share training content. The six test pairs sample training randomness within one fixed class partition.

MobileNet uses FP32 throughout content fitting and test runs. Other models use AMP, and Mixer uses activation recomputation. Precision choices are uniform across the two branches of each model condition.

\subsection{Complete cross-coverage and bilateral contrasts}
Both branch contrasts have positive mean advantages and positive 95\% paired-seed intervals for every model on the first common task (Figure~\ref{fig:content}), giving 16 positive branch-level intervals across the eight model--dataset conditions. On CIFAR-10's first common task, ResNet's own-history advantages are 3.99 and 5.51 percentage points; DeiT's are 7.14 and 8.84 points. The second common task on CIFAR-10 supplies eight further positive branch-level intervals (Figure~\ref{fig:content_second}), for 24 across all measured common tasks.

Table~\ref{tab:cross} reports the four cells $C_{h\leftarrow s}=\covg(D_h,U_s)$; Table~\ref{tab:content} reports their paired differences. Difference intervals are computed directly from paired observations. Every branch-level mean contrast is positive with a positive nominal 95\% interval.

\begin{table}[htbp]
\centering\small
\caption{Complete cross-coverage means (\%) during common tasks. Rows of each $2\times2$ comparison correspond to the test branch and columns to the scaffold's early history.}
\label{tab:cross}
\begin{tabular}{llrrrrr}\toprule
Model & CIFAR & Task & $X\leftarrow X$ & $X\leftarrow Y$ & $Y\leftarrow X$ & $Y\leftarrow Y$\\\midrule
ResNet-18 & 10 & 3 & 26.85 & 22.86 & 20.88 & 26.38 \\
ResNet-18 & 10 & 4 & 22.87 & 20.76 & 20.17 & 24.23 \\
DeiT-Tiny & 10 & 3 & 29.55 & 22.41 & 22.37 & 31.21 \\
DeiT-Tiny & 10 & 4 & 18.94 & 18.26 & 19.08 & 21.65 \\
MobileNetV3-S & 10 & 3 & 13.71 & 11.90 & 14.06 & 15.96 \\
MobileNetV3-S & 10 & 4 & 11.88 & 11.15 & 12.93 & 14.17 \\
MLP-Mixer-B/16 & 10 & 3 & 22.35 & 18.72 & 21.71 & 25.37 \\
MLP-Mixer-B/16 & 10 & 4 & 20.81 & 19.33 & 18.58 & 20.42 \\
ResNet-18 & 100 & 3 & 17.01 & 13.37 & 11.61 & 16.97 \\
DeiT-Tiny & 100 & 3 & 38.41 & 22.44 & 22.65 & 37.73 \\
MobileNetV3-S & 100 & 3 & 9.05 & 6.06 & 5.34 & 9.93 \\
MLP-Mixer-B/16 & 100 & 3 & 17.29 & 13.64 & 12.92 & 16.81 \\
\bottomrule

\end{tabular}
\end{table}

\begin{table}[htbp]
\centering\small
\caption{Own-history coverage advantages in percentage points [95\% paired-seed interval]. Each condition uses six separate test pairs, resampled jointly 10,000 times, conditional on the fitted scaffolds.}
\label{tab:content}
\begin{tabular}{llrrr}\toprule
Model & CIFAR & Task & $\Delta_X$ & $\Delta_Y$\\\midrule
ResNet-18 & 10 & 3 & 3.99 [3.51, 4.54] & 5.51 [5.01, 5.99] \\
ResNet-18 & 10 & 4 & 2.12 [1.53, 2.70] & 4.06 [3.74, 4.38] \\
DeiT-Tiny & 10 & 3 & 7.14 [4.83, 9.75] & 8.84 [7.36, 10.62] \\
DeiT-Tiny & 10 & 4 & 0.68 [0.22, 1.18] & 2.58 [2.01, 3.21] \\
MobileNetV3-S & 10 & 3 & 1.81 [1.31, 2.45] & 1.90 [1.48, 2.33] \\
MobileNetV3-S & 10 & 4 & 0.73 [0.14, 1.30] & 1.24 [1.04, 1.44] \\
MLP-Mixer-B/16 & 10 & 3 & 3.62 [1.60, 4.81] & 3.66 [1.30, 6.05] \\
MLP-Mixer-B/16 & 10 & 4 & 1.48 [0.12, 2.64] & 1.83 [1.20, 2.53] \\
ResNet-18 & 100 & 3 & 3.64 [3.04, 4.27] & 5.36 [4.93, 5.79] \\
DeiT-Tiny & 100 & 3 & 15.98 [15.17, 16.75] & 15.09 [14.26, 15.84] \\
MobileNetV3-S & 100 & 3 & 2.98 [2.39, 3.61] & 4.59 [4.24, 5.03] \\
MLP-Mixer-B/16 & 100 & 3 & 3.65 [3.17, 4.18] & 3.89 [3.45, 4.41] \\
\bottomrule

\end{tabular}
\end{table}
\FloatBarrier

The ten-class tasks on CIFAR-100 give the same qualitative result. ResNet's branch advantages are 3.64 and 5.36 points, MobileNet's 2.98 and 4.59, and Mixer's 3.65 and 3.89. DeiT has larger advantages of 15.98 and 15.09 points in this condition. Individual histories also vary: on CIFAR-10, Mixer's paired average is positive in five of six test seeds at each common task, although both branch-mean intervals remain positive.

Scaffold overlap is $\|U_X^\top U_Y\|_F^2/r$. The point estimates on CIFAR-10 are 57.46\%, 58.72\%, 44.19\%, and 50.12\% for ResNet, DeiT, MobileNet, and Mixer; corresponding CIFAR-100 overlaps are 26.62\%, 47.21\%, 14.24\%, and 58.92\%. These orientation summaries complement the held-out bilateral coverage contrasts.

\subsection{Persistence through a second common task}
\begin{figure}[htbp]
\centering
\includegraphics[width=.82\linewidth]{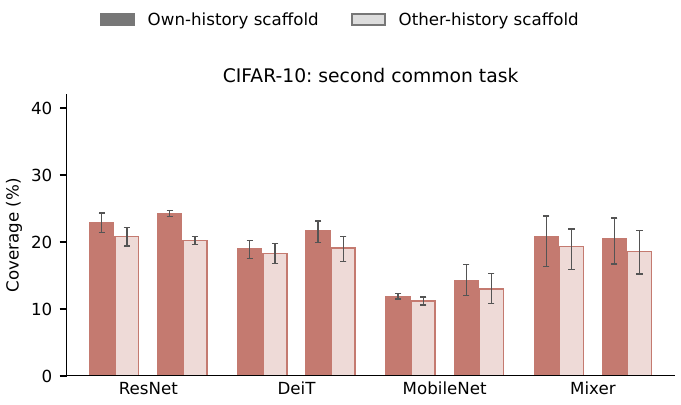}
\caption{\textbf{History-specific scaffold preference persists through a second common task.} Absolute coverage on CIFAR-10 task four. For each model, the left pair of bars evaluates test branch $X$ and the right pair evaluates test branch $Y$; within each pair, the solid left bar uses the own-history scaffold and the pale right bar uses the other-history scaffold. Bars average six paired test seeds. Error bars are nominal 95\% intervals from 10,000 paired-seed bootstrap draws, conditional on the fitted scaffolds. Table~\ref{tab:content} reports the corresponding paired contrasts.}
\label{fig:content_second}
\end{figure}

CIFAR-10 branches complete a second common task after training on the first common task with matched inputs. During this follow-up, both branches replay only task $A$ and the preceding common task; distinctive $B_X$ and $B_Y$ images remain excluded. Later drift still favors each branch's own-history scaffold in all four models (Figure~\ref{fig:content_second}). All eight paired own-minus-cross intervals are positive. For MobileNet, the two advantages are 0.73 and 1.24 percentage points. This follow-up tests whether the historical preference survives an additional stage of shared training.

\FloatBarrier

\section{Response Geometry and Actual Updates}\label{app:response}
\subsection{Local parameter response}
Collect the $p$ encoder parameters in $\theta$. To obtain a random direction, fill $g_j$ with independent standard normal values and divide it by its Euclidean length $\|g_j\|_2$. Set $\delta\theta_j=\varepsilon g_j/\|g_j\|_2$, with length $\varepsilon=10^{-4}\|\theta\|_2$. With classifier weights and normalization buffers fixed, define
\begin{equation}
L_j=\tfrac12H[Z_{\theta+\delta\theta_j}-Z_{\theta-\delta\theta_j}],\qquad
\widehat B_\theta=\frac1K\sum_{j=1}^{K}L_j^\top L_j.
\label{eq:response_main}
\end{equation}
Here $K$ is the number of perturbations, sixteen in our experiments, and $Z_{\theta\pm\delta\theta_j}$ contains the anchor features after adding or subtracting the same small parameter change. The symmetric difference $L_j$ measures the corresponding centered feature response. The $d\times d$ matrix $\widehat B_\theta$ collects average response energy across feature directions; its leading $r$ eigenvectors maximize retained response energy. We pool their projectors across the six estimation histories at each endpoint, using the scaffold's fixed rank.

The response measurement uses the symmetric finite differences in Equation~\ref{eq:response_main}. Encoder parameters are perturbed in sixteen independent random directions of equal length, with classifier weights and stored normalization statistics fixed. Perturbation directions and anchors are reused across checkpoints; perturbation length scales with the encoder parameter norm. Response projectors are pooled across six estimation histories at each checkpoint.

The connection to parameter sensitivity can be expressed using a local linear map. Let $G_i\in\R^{d\times p}$ map a small change in the $p$ encoder parameters to the change in input $i$'s centered feature vector. This matrix is the \emph{Jacobian}: entry $(a,b)$ is the rate of change of centered feature $a$ when parameter $b$ changes. Near a differentiable network state, the response is approximately $G_i\delta\theta$. For a uniformly oriented parameter change of length $\varepsilon$, the average outer product is $(\varepsilon^2/p)I_p$, because each of the $p$ coordinates receives the same expected squared change. Therefore
\begin{equation}
B_{\rm iso}\simeq\frac{\varepsilon^2}{p}\sum_{i=1}^nG_iG_i^\top,
\qquad A_v^\top A_v\simeq\sum_{i=1}^nG_i vv^\top G_i^\top.
\label{eq:update_bridge}
\end{equation}
Here $B_{\rm iso}$ is the average of $L^\top L$ over isotropic perturbations, $v$ is a particular optimizer parameter step, and $A_v=H(Z_{\theta+v}-Z_\theta)$ is its centered feature motion with normalization buffers fixed. The symbol $\simeq$ indicates the local linear approximation. Both matrices are $d\times d$ and summarize feature-direction energy over inputs. The first expression weights parameter directions equally; the second weights the direction selected by training. Their leading feature directions can consequently differ. We measure these responses directly from finite feature differences.

Overlap gives each retained basis direction equal weight. By contrast, the fraction of average response energy captured by a scaffold is $\operatorname{tr}(\Usc^\top B_{\rm iso}\Usc)/\operatorname{tr}(B_{\rm iso})$, which also weights directions by their response magnitude. This explains why orientation overlap and motion coverage answer different questions.

\subsection{Multi-position measurements}\label{app:updates}
The replay update experiment resumes each saved pre-final-task network with a fresh task-boundary AdamW optimizer and deterministic augmentation seeds. CIFAR-10 windows use successful steps 2--4, 65--67, and 125--127 within 145 planned attempts; CIFAR-100 uses 2--4, 316--318, and 608--610 within 705 planned attempts. Overflow-skipped attempts do not count as successful steps.

For each update, full-state feature motion compares model states immediately before and after training. The fixed-buffer field uses post-update parameters with pre-update normalization buffers. A third field applies the captured training gradient after matching its encoder norm to the actual parameter displacement. All fields use deterministic FP32 evaluation on fixed held-out anchors and are centered over inputs. Each coverage is normalized by its own field energy. The classifier-matched reference uses the pre-final-task old-class classifier throughout.

Three steps are averaged within each history and window before hierarchical resampling of six orders and two runs per order. There are 96 test histories, 288 history-window means, and 864 sampled AdamW steps. Five step gains are nonpositive: one in DeiT/CIFAR-10 and four in Mixer/CIFAR-10. One history-window mean is nonpositive. All twenty-four mean-window intervals and all twenty-four norm-matched-gradient intervals are positive. Table~\ref{tab:updates} reports the primary comparisons.

All 24 condition-by-window mean gains have positive nominal 95\% intervals (Figure~\ref{fig:updates}). On CIFAR-10, ResNet's early, middle, and late gains are 9.23, 7.92, and 8.04 points. Mixer's are 2.82, 3.20, and 3.90 points. On CIFAR-100, DeiT's gains are 1.68, 1.46, and 1.36 points, while Mixer's are 2.93, 3.45, and 3.82 points. The early subspace therefore captures a preferred component of motion throughout the sampled course of the task.

\begin{table}[htbp]
\centering\small
\caption{Actual AdamW coverage gains over classifier-matched random subspaces (percentage points [95\% hierarchical interval]). Windows average three neighboring updates within each history.}
\label{tab:updates}
\begin{tabular}{llrrr}\toprule
Model & CIFAR & Early & Middle & Late\\\midrule
ResNet-18 & 10 & 9.23 [7.14, 10.88] & 7.92 [6.04, 9.43] & 8.04 [6.04, 9.57] \\
ResNet-18 & 100 & 8.02 [7.08, 9.17] & 5.50 [5.19, 5.83] & 6.17 [5.84, 6.54] \\
DeiT-Tiny & 10 & 1.34 [0.78, 2.02] & 1.20 [0.83, 1.63] & 1.07 [0.73, 1.45] \\
DeiT-Tiny & 100 & 1.68 [1.35, 2.05] & 1.46 [1.17, 1.74] & 1.36 [0.92, 1.92] \\
MobileNetV3-S & 10 & 5.18 [4.36, 6.17] & 4.12 [3.80, 4.46] & 4.10 [3.79, 4.45] \\
MobileNetV3-S & 100 & 1.85 [1.71, 2.00] & 1.84 [1.67, 2.05] & 1.85 [1.69, 2.02] \\
MLP-Mixer-B/16 & 10 & 2.82 [1.76, 3.93] & 3.20 [1.70, 4.71] & 3.90 [2.51, 5.38] \\
MLP-Mixer-B/16 & 100 & 2.93 [2.35, 3.46] & 3.45 [2.76, 4.19] & 3.82 [3.11, 5.06] \\
\bottomrule

\end{tabular}
\end{table}

\section{Frozen-Classifier Consequences}\label{app:readout}
The readout analysis uses the baseline final-task displacement and the old-class classifier at its pre-final-task state: eight classes on CIFAR-10 and thirty on CIFAR-100. These endpoint measurements complement the update continuations in Section~\ref{sec:updates}.

The scaffold also captures feature motion that can affect class scores. Holding the pre-final-task classifier fixed, we reconstruct its changes in logits (class scores before normalization into probabilities) and cross-entropy loss from the projected feature displacement. Both the scaffold and matched random reconstruction receive the same mean feature displacement, so the comparison concerns the centered component defined in Equation~\ref{eq:drift}. For a measured change $y$ and its reconstruction $\hat y$, the score $1-\|y-\hat y\|^2/\|y\|^2$ quantifies fidelity: higher scores indicate closer reconstruction, and negative scores indicate error energy exceeding the full change energy.

Write the full displacement as $\Delta Z=D+\mathbf1_n\mu^\top$, with $\mu=\Delta Z^\top\mathbf1_n/n$, and supply both reconstructions with the translated baseline $Z_*=Z_{\rm pre}+\mathbf1_n\mu^\top$. For fixed classifier weights $W\in\R^{c\times d}$, let $H_c=I_c-\mathbf1_c\mathbf1_c^\top/c$ center scores across classes. The full and projected logit changes are
\begin{equation}
Y=DW^\top H_c,\qquad \widehat Y_U=DUU^\top W^\top H_c,\qquad
R_{\rm logit}(U)=1-\frac{\|Y-\widehat Y_U\|_F^2}{\|Y\|_F^2}.
\label{eq:readout_main}
\end{equation}
Classifier bias cancels in these differences. The analogous loss score reconstructs per-input cross-entropy changes from $Z_*$. Both scores test the functional visibility of centered motion, conditional on its supplied mean displacement.

For logits, subtract the class-wise mean for each input. For loss, evaluate cross-entropy against the true old-class label before and after adding the centered displacement. If $y$ is the full change and $\hat y$ its reconstruction, define
\begin{equation}
R=1-\frac{\|y-\hat y\|^2}{\|y\|^2}.
\end{equation}
The norm is Frobenius for the logit matrix and Euclidean for the loss-change vector. The score can be negative. Supplying the mean displacement conditions reconstruction on that shared component. A positive gain denotes more accurate reconstruction of the centered output change than the matched random reference.

Output effects are more condition-dependent than geometric recurrence. On CIFAR-100, Mixer has reconstruction scores of 25.70\% for centered logit changes and 22.66\% for loss changes, improving over the matched control by 10.12 and 4.41 points. Its CIFAR-10 gains are 5.68 and 4.71 points; both intervals are positive on both datasets. DeiT's CIFAR-10 logit gain, ResNet's CIFAR-10 loss gain, and MobileNet's CIFAR-100 loss gain have intervals spanning zero. Table~\ref{tab:readout} gives all condition-level contrasts. Recurring feature motion therefore reaches the fixed classifier, with reconstruction gains that vary across conditions.

\begin{table}[htbp]
\centering\small
\caption{Frozen-classifier reconstruction gains over the matched random control (percentage points [95\% paired hierarchical interval]). These baseline endpoint measurements are distinct from the multi-position update continuations.}
\label{tab:readout}
\begin{tabular}{llrr}\toprule
Model & CIFAR & Class-centered logits & Loss change\\\midrule
ResNet-18 & 10 & 6.09 [3.06, 8.82] & 7.12 [-1.59, 15.45] \\
ResNet-18 & 100 & 13.48 [12.14, 15.05] & 17.44 [14.93, 20.41] \\
DeiT-Tiny & 10 & 0.83 [-0.01, 1.66] & 1.81 [0.01, 3.75] \\
DeiT-Tiny & 100 & 1.50 [0.47, 2.42] & 1.30 [0.03, 2.73] \\
MobileNetV3-S & 10 & 3.89 [1.59, 6.34] & 6.71 [3.56, 10.09] \\
MobileNetV3-S & 100 & 1.13 [0.62, 1.69] & 0.54 [-0.72, 2.06] \\
MLP-Mixer-B/16 & 10 & 5.68 [2.53, 8.51] & 4.71 [0.27, 9.54] \\
MLP-Mixer-B/16 & 100 & 10.12 [8.78, 11.59] & 4.41 [2.60, 6.09] \\
\bottomrule

\end{tabular}
\end{table}
\FloatBarrier

\section{Recurrence Without Replay}\label{app:noreplay}
Each model learns five two-class CIFAR-10 tasks without replay, using 450 training images per class and five epochs. Batches contain 64 current images, giving 75 successful updates per task. Each model's full initial state is identical across its no-replay histories. The estimation/test order identifiers match the baseline CIFAR-10 sets, and scaffold ranks are carried over from the baseline selection.

Scaffolds are fitted separately from six no-replay estimation histories' $1\to2$ drift on 500 first-task training anchors. Twelve other histories are evaluated on 100 excluded anchors. These anchors are measured after learning but are never used as old-example training replay. No-replay training uses FP32 for MobileNet and AMP for the other three models. Mixer uses activation recomputation. The response analysis uses sixteen paired FP32 parameter perturbations per endpoint.

Classifier-matched references use transition-start classifiers and exact expected projection energy. Matched-control contrasts vary, including the negative final Mixer result in Table~\ref{tab:noreplay}. Actual-update observations use steps 2, 38, and 73 in each of tasks two through five: 144 observations per model, grouped into twelve task-by-position comparisons over twelve histories. Positive gain intervals occur in twelve comparisons for ResNet, four for DeiT, twelve for MobileNet, and three for Mixer. Individual gains are positive in 144, 103, 144, and 77 observations, respectively.

\begin{figure}[htbp]
\centering
\includegraphics[width=.9\linewidth]{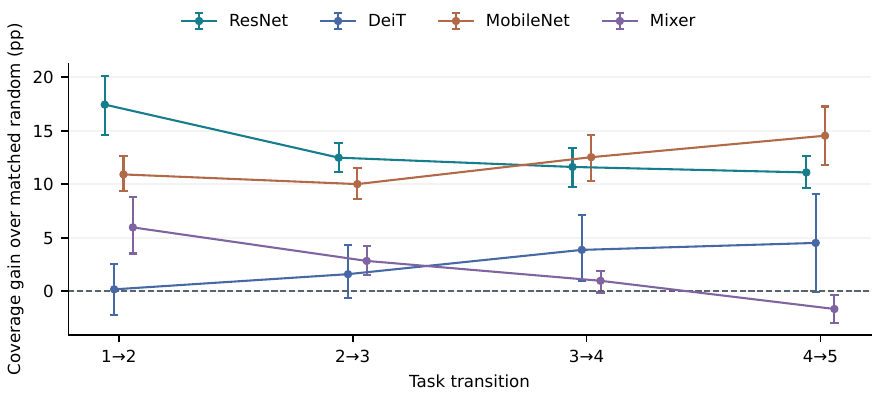}
\caption{\textbf{Recurrence without replay varies across conditions.} Fixed-scaffold coverage minus the transition-start classifier-matched random expectation. Error bars are nominal 95\% hierarchical order/run bootstrap intervals over twelve histories, conditional on each fitted scaffold.}
\label{fig:noreplay}
\end{figure}

\begin{table}[htbp]
\centering\small
\caption{Complete no-replay transition gains (percentage points [95\% hierarchical interval]). The matched-control gain uses total centered drift energy.}
\label{tab:noreplay}
\begin{tabular}{llr}\toprule
Model & Transition & Matched-control gain\\\midrule
ResNet-18 & $1\to 2$ & 17.44 [14.63, 20.11] \\
ResNet-18 & $2\to 3$ & 12.48 [11.15, 13.83] \\
ResNet-18 & $3\to 4$ & 11.61 [9.76, 13.38] \\
ResNet-18 & $4\to 5$ & 11.10 [9.63, 12.65] \\
DeiT-Tiny & $1\to 2$ & 0.17 [-2.20, 2.55] \\
DeiT-Tiny & $2\to 3$ & 1.59 [-0.68, 4.35] \\
DeiT-Tiny & $3\to 4$ & 3.86 [0.99, 7.11] \\
DeiT-Tiny & $4\to 5$ & 4.51 [-0.09, 9.09] \\
MobileNetV3-S & $1\to 2$ & 10.92 [9.39, 12.63] \\
MobileNetV3-S & $2\to 3$ & 10.01 [8.64, 11.48] \\
MobileNetV3-S & $3\to 4$ & 12.53 [10.27, 14.63] \\
MobileNetV3-S & $4\to 5$ & 14.53 [11.82, 17.26] \\
MLP-Mixer-B/16 & $1\to 2$ & 5.96 [3.52, 8.84] \\
MLP-Mixer-B/16 & $2\to 3$ & 2.83 [1.49, 4.20] \\
MLP-Mixer-B/16 & $3\to 4$ & 0.97 [-0.12, 1.91] \\
MLP-Mixer-B/16 & $4\to 5$ & -1.66 [-2.96, -0.37] \\
\bottomrule

\end{tabular}
\end{table}
\FloatBarrier

The no-replay study evaluates recurrence under a second training regime, with its own initialization arrangement and update budget. It measures scaffold recurrence under that learning regime.

\section{Direction-Suppression Intervention}\label{app:intervention}
\subsection{Paired design and outcomes}
Each of the eight model--dataset conditions uses six histories: three task orders with two runs each. CIFAR-10 uses orders 2131, 2137, and 2141 with seeds 170 and 171; CIFAR-100 uses orders 1123, 1129, and 1151 with seeds 150 and 151. The fixed-coefficient experiments use $\lambda=10$ and 100 for the scaffold and two uniform-random constraints, alongside normal training. Table~\ref{tab:intervention} shows the coefficient-100 results with both uniform-random and energy-matched controls. Ranks are those in Appendix~\ref{app:rank}. Uniform random bases use seeds 91001 and 91002 and remain fixed across histories within each model condition.

Every branch starts from the same pre-final-task state within its history and uses a fresh AdamW optimizer. Training uses learning rate $3\times10^{-4}$, weight decay $0.05$, five epochs, batches of 32 current and 32 replay examples, and 500 old training-memory examples. Paired branches receive the same augmented current/replay stream. The penalty in Equation~\ref{eq:suppress} uses deterministic FP32 evaluation-mode features, with gradients updating encoder parameters. MobileNet classification uses FP32; the other models use AMP. Each branch completes 145 updates on CIFAR-10 or 705 on CIFAR-100. Within each model-dataset condition, scaffold and random branches use the same rank and penalty normalization. Each comparison therefore measures acquisition and retention under a common quadratic penalty applied to different feature subspaces.

Evaluation uses the excluded last fifty training-split images per class. New-task accuracy measures acquisition of the final task. First-task accuracy measures retention of the initial task under competition among all seen classes: ten on CIFAR-10 and forty on CIFAR-100. For branch $b$ in history $h$, let $a_{h,b,e}^{\rm new}$ be new-task accuracy after epoch $e$. With $E$ training epochs, define
\begin{equation}
\overline a_{h,b}^{\rm new}=\frac1E\sum_{e=1}^E a_{h,b,e}^{\rm new},\qquad
\Delta_h^{\rm new}=\overline a_{h,\rm sc}^{\rm new}
-\frac{\overline a_{h,\rm rand1}^{\rm new}+\overline a_{h,\rm rand2}^{\rm new}}2.
\end{equation}
Here $E$ is five; the pretraining epoch-zero measurement is excluded. First-task contrasts use the final epoch's first-task accuracy. All-old-task retention averages final accuracy across preceding tasks. Comparisons with normal training subtract the corresponding value. Random-branch accuracies are averaged within each history before computing the paired contrasts. Means give equal weight to orders. Nominal 95\% intervals use 20,000 hierarchical bootstrap draws, resampling the three orders and then the two runs within each selected order, with the scaffold fixed. The training history is the unit of replication; epochs, images, and random branches are summarized within each history.

\subsection{Acquisition and retention at two penalty strengths}
Tables~\ref{tab:suppression10} and~\ref{tab:suppression100} report acquisition and retention at $\lambda=10$ and 100. At $\lambda=100$, mean acquisition is lower than under random constraints in all eight conditions, with intervals below zero in five. Seven conditions also have intervals below zero relative to normal training; MobileNet/CIFAR-10 is the exception. Mixer/CIFAR-100 has a positive first-task retention interval relative to both controls; the other seven first-task contrasts against random include zero at the reported precision. At this coefficient, the acquisition cost is more consistent across conditions than a loss of old-task retention.

Mixer/CIFAR-100 illustrates the different acquisition and retention outcomes at $\lambda=100$. First-task retention improves by 3.03 points [0.53, 5.33] relative to normal training. Mean acquisition falls by 5.61 points [4.56, 6.61]. Thus, an acquisition cost can accompany a retention benefit.

\begin{table}[htbp]
\centering\scriptsize
\caption{CIFAR-10 suppression at $\lambda=10$ and 100: scaffold minus uniform random (R) or normal training (B), in percentage points [95\% paired hierarchical interval]. New-task accuracy averages five epochs; first-task accuracy is final.}
\label{tab:suppression10}
\setlength{\tabcolsep}{3pt}
\begin{tabular}{lrllll}\toprule
 & & \multicolumn{2}{c}{New-task acquisition} & \multicolumn{2}{c}{First-task retention}\\
Model & $\lambda$ & versus R & versus B & versus R & versus B\\\midrule
ResNet-18 & 10 & -11.78 [-13.63, -8.83] & -13.63 [-17.40, -9.07] & +0.67 [-1.08, +3.08] & +0.83 [-3.33, +4.50]\\
ResNet-18 & 100 & -6.97 [-8.15, -5.80] & -20.93 [-24.53, -16.20] & +1.67 [-1.00, +4.83] & +0.33 [-3.00, +2.50]\\
DeiT-Tiny & 10 & -0.62 [-2.15, +1.00] & -3.03 [-5.57, +0.40] & +0.83 [-2.50, +4.50] & +4.00 [-2.67, +10.00]\\
DeiT-Tiny & 100 & -0.35 [-2.93, +1.47] & -19.03 [-23.10, -14.70] & +1.50 [-2.50, +4.50] & +1.67 [-7.67, +8.50]\\
MobileNetV3-S & 10 & -0.40 [-2.03, +1.20] & -1.30 [-14.67, +5.07] & -0.08 [-4.92, +5.42] & -1.67 [-18.67, +12.33]\\
MobileNetV3-S & 100 & -4.75 [-12.40, +0.10] & -8.37 [-29.87, +2.93] & +2.67 [-1.17, +6.00] & -1.17 [-13.33, +13.50]\\
MLP-Mixer-B/16 & 10 & -8.13 [-11.28, -2.33] & -10.70 [-14.73, -6.50] & +4.67 [-0.25, +9.00] & +3.17 [-5.50, +11.50]\\
MLP-Mixer-B/16 & 100 & -8.32 [-12.85, -3.15] & -17.27 [-22.07, -10.90] & +1.08 [-6.00, +6.17] & +2.00 [-12.50, +13.00]\\
\bottomrule

\end{tabular}
\end{table}
\begin{table}[htbp]
\centering\scriptsize
\caption{CIFAR-100 suppression at $\lambda=10$ and 100: paired acquisition and retention differences, using the definitions in Table~\ref{tab:suppression10}.}
\label{tab:suppression100}
\setlength{\tabcolsep}{3pt}
\begin{tabular}{lrllll}\toprule
 & & \multicolumn{2}{c}{New-task acquisition} & \multicolumn{2}{c}{First-task retention}\\
Model & $\lambda$ & versus R & versus B & versus R & versus B\\\midrule
ResNet-18 & 10 & -6.74 [-7.25, -6.23] & -7.37 [-7.96, -6.79] & -0.37 [-1.90, +1.37] & -0.40 [-1.80, +1.73]\\
ResNet-18 & 100 & -6.07 [-7.00, -5.21] & -14.11 [-15.11, -13.29] & +0.17 [-0.85, +1.23] & -1.77 [-4.13, +0.70]\\
DeiT-Tiny & 10 & -0.43 [-0.65, -0.01] & +1.38 [+0.01, +2.27] & -0.33 [-1.65, +1.15] & +4.43 [+0.57, +8.80]\\
DeiT-Tiny & 100 & -0.79 [-1.36, -0.28] & -6.75 [-8.09, -5.65] & -0.13 [-1.10, +0.77] & +3.00 [-0.33, +6.37]\\
MobileNetV3-S & 10 & -1.30 [-2.64, -0.17] & -2.60 [-5.09, -0.94] & +1.88 [-1.03, +5.32] & +2.47 [-4.07, +8.97]\\
MobileNetV3-S & 100 & -0.87 [-2.42, +1.74] & -10.71 [-12.68, -9.11] & +1.85 [+0.00, +4.27] & +6.27 [+0.27, +12.27]\\
MLP-Mixer-B/16 & 10 & -2.29 [-3.58, -1.40] & +0.28 [-0.58, +1.28] & +6.33 [+3.90, +8.98] & +3.20 [+1.07, +5.50]\\
MLP-Mixer-B/16 & 100 & -3.49 [-4.31, -2.83] & -5.61 [-6.61, -4.56] & +3.30 [+1.40, +5.45] & +3.03 [+0.53, +5.33]\\
\bottomrule

\end{tabular}
\end{table}
\FloatBarrier

\subsection{Energy-matched random constraints}\label{app:energy_matched}
The energy-matched comparison uses the same 48 histories, normal-training branches, and scaffold branches at $\lambda=100$. Within each history, we construct two random bases from the final-task drift of the 500 old training-memory examples under normal training. Each basis has the scaffold's rank and captures the same drift energy. The bases remain fixed during the subsequent constrained continuations.

Let $D_b^{\rm mem}$ be the centered feature displacement of these memory examples from the pre-final-task state to the final epoch of branch $b$. Write $b=0$ for normal training and define the energy along a basis $U$ as
\begin{equation}
E_b(U)=\frac{1}{N}\|D_b^{\rm mem}U\|_F^2,
\qquad N=500.
\end{equation}
We form $C=(D_0^{\rm mem})^\top D_0^{\rm mem}/N$ and decompose it as $C=Q\operatorname{diag}(\nu_i)Q^\top$. Starting from a uniform random orthonormal basis $G$, we obtain an energy-matched basis by orthonormalizing
\begin{equation}
Q\operatorname{diag}(s_i^\alpha)Q^\top G,
\qquad s_i=\max(\nu_i/\nu_{\max},10^{-12}).
\end{equation}
We choose $\alpha$ by bisection so that $E_0(U_{\rm rand})=E_0(\Usc)$ within relative tolerance $10^{-5}$. The two starting bases use seeds 91001 and 91002. This procedure fits the random bases separately within each history using the normal-training memory drift.

We next adjust the penalty coefficient for each random basis. The scaffold branch supplies the reference reduction $R_{\rm sc}=E_0(\Usc)-E_{\rm sc}(\Usc)$. For each candidate random coefficient, we run a paired five-epoch continuation and measure $R_{\rm rand}=E_0(U_{\rm rand})-E_{\rm rand}(U_{\rm rand})$. We select a coefficient satisfying $|R_{\rm rand}-R_{\rm sc}|/E_0(U_{\rm rand})\leq0.01$. Coefficient selection uses memory-feature energy. All 96 random branches meet this criterion; the largest discrepancy is 0.96 percentage points of their baseline projected energy. Table~\ref{tab:energy_matching} reports the matching results.

\begin{table}[htbp]
\centering\small
\caption{Matching drift-energy reductions on old training-memory examples. Reduction is $100[E_0(U)-E_b(U)]/E_0(U)$. Means average six histories and, for random constraints, two bases per history. Maximum error is the largest matching discrepancy across individual random branches.}
\label{tab:energy_matching}
\setlength{\tabcolsep}{5pt}
\begin{tabular}{lr rrr l}\toprule
 & & \multicolumn{2}{c}{Energy reduction (\%)} & & \\
Model & CIFAR & Scaffold & Random & Max. error (pp) & Random $\lambda$\\\midrule
ResNet-18 & 10 & 89.49 & 89.74 & 0.92 & 0.21--0.87\\
ResNet-18 & 100 & 95.33 & 95.69 & 0.88 & 0.21--1.00\\
DeiT-Tiny & 10 & 97.59 & 97.61 & 0.73 & 10.00--100.00\\
DeiT-Tiny & 100 & 98.87 & 98.51 & 0.87 & 10.00\\
MobileNetV3-S & 10 & 94.35 & 94.33 & 0.96 & 6.50--36.26\\
MobileNetV3-S & 100 & 98.54 & 98.08 & 0.84 & 1.00--100.00\\
MLP-Mixer-B/16 & 10 & 96.50 & 96.72 & 0.90 & 0.69--100.00\\
MLP-Mixer-B/16 & 100 & 98.67 & 98.66 & 0.53 & 10.00\\
\bottomrule

\end{tabular}
\end{table}

Acquisition and retention use the same epoch summaries as the fixed-coefficient experiments. Random-branch accuracies are averaged within history. Intervals resample paired orders and runs, keeping the fitted bases and selected coefficients fixed. All eight acquisition intervals against energy-matched random constraints lie below zero (Table~\ref{tab:energy_matched}). For all-old-task retention, ResNet/CIFAR-100 and Mixer/CIFAR-100 have positive intervals, DeiT/CIFAR-10 has a negative interval, and the other five span zero. First-task retention has a positive interval for MobileNet/CIFAR-100 and a negative interval for DeiT/CIFAR-100; the other six span zero.

\begin{table}[htbp]
\centering\scriptsize
\caption{Energy-matched intervention: scaffold minus random (R) or normal training (B), in percentage points [95\% paired hierarchical interval]. Acquisition averages five epochs; retention uses the final epoch. The scaffold coefficient is 100.}
\label{tab:energy_matched}
\setlength{\tabcolsep}{3pt}
\begin{tabular}{lrllll}\toprule
 & & \multicolumn{2}{c}{New-task acquisition} & First-task retention & All-old-task retention\\
Model & CIFAR & versus R & versus B & versus R & versus R\\\midrule
ResNet-18 & 10 & -21.58 [-25.15, -17.75] & -20.93 [-24.53, -16.20] & -1.75 [-4.83, +0.17] & -2.46 [-5.63, +0.67]\\
ResNet-18 & 100 & -13.90 [-14.85, -13.05] & -14.11 [-15.11, -13.29] & -1.07 [-3.70, +1.57] & +1.79 [+0.39, +3.40]\\
DeiT-Tiny & 10 & -11.72 [-17.43, -4.25] & -19.03 [-23.10, -14.70] & -0.50 [-4.58, +3.50] & -2.83 [-5.08, -0.85]\\
DeiT-Tiny & 100 & -8.37 [-8.95, -7.63] & -6.75 [-8.09, -5.65] & -1.75 [-3.92, -0.03] & +0.51 [-0.64, +1.72]\\
MobileNetV3-S & 10 & -6.47 [-14.95, -0.50] & -8.37 [-29.87, +2.93] & +1.25 [-4.00, +7.67] & +0.67 [-2.75, +3.21]\\
MobileNetV3-S & 100 & -6.76 [-9.98, -2.53] & -10.71 [-12.68, -9.11] & +2.67 [+0.70, +6.63] & +0.63 [-0.90, +2.76]\\
MLP-Mixer-B/16 & 10 & -8.33 [-14.90, -0.90] & -17.27 [-22.07, -10.90] & +0.33 [-6.33, +5.33] & -1.75 [-6.17, +1.79]\\
MLP-Mixer-B/16 & 100 & -7.17 [-7.90, -6.36] & -5.61 [-6.61, -4.56] & +0.55 [-1.15, +2.03] & +1.89 [+0.93, +3.08]\\
\bottomrule

\end{tabular}
\end{table}
\FloatBarrier

\section{Complete Task-Accuracy Results}\label{app:accuracy}
All entries are percentages on held-out images excluded from training and replay. Task numbers denote positions within a history. Predictions compete over the classes available at that checkpoint, without a task identifier. The tables give absolute accuracies; the paired effect estimates are reported in Appendix~\ref{app:intervention}.
\subsection{Ordinary continual learning}
Tables~\ref{tab:accuracy_base10} and~\ref{tab:accuracy_base100} show accuracy on every learned task after each training task. Entries average twelve test histories unless a superscript gives a smaller count. A dash marks a task not yet learned; NR marks an unavailable measurement. These histories are separate from the six paired intervention histories.
\begingroup\small\setlength{\tabcolsep}{4pt}\setlength{\LTpre}{6pt}\setlength{\LTpost}{6pt}
\begin{longtable}{llrrrrr}
\caption{CIFAR-10 baseline task-accuracy matrix. Rows indicate the task just completed; columns indicate the evaluated task.}\label{tab:accuracy_base10}\\
\toprule
Model & After task & T1 & T2 & T3 & T4 & T5\\\midrule
\endfirsthead
\multicolumn{7}{l}{\tablename\ \thetable\ (continued)}\\
\toprule
Model & After task & T1 & T2 & T3 & T4 & T5\\\midrule
\endhead
\midrule\multicolumn{7}{r}{Continued on next page}\\\endfoot
\bottomrule\endlastfoot
ResNet-18 & 1 & 94.00 & -- & -- & -- & --\\
 & 2 & 88.42 & 95.17 & -- & -- & --\\
 & 3 & 82.25 & 84.42 & 94.58 & -- & --\\
 & 4 & 78.08 & 79.58 & 75.58 & 93.83 & --\\
 & 5 & 69.33 & 74.33 & 74.83 & 77.75 & 93.42\\
\addlinespace
DeiT-Tiny & 1 & 89.08 & -- & -- & -- & --\\
 & 2 & 85.83 & 91.30$^{10}$ & -- & -- & --\\
 & 3 & 78.58 & 84.80$^{10}$ & 92.50$^{10}$ & -- & --\\
 & 4 & 74.83 & 79.58 & 73.25 & 91.58 & --\\
 & 5 & 67.08 & 72.17 & 66.50 & 77.42 & 90.67\\
\addlinespace
MobileNetV3-S & 1 & 79.75 & -- & -- & -- & --\\
 & 2 & 82.42 & NR & -- & -- & --\\
 & 3 & 68.83 & NR & NR & -- & --\\
 & 4 & 65.08 & 67.17 & 63.50 & 85.50 & --\\
 & 5 & 59.75 & 62.33 & 58.58 & 65.25 & 86.58\\
\addlinespace
MLP-Mixer-B/16 & 1 & 84.92 & -- & -- & -- & --\\
 & 2 & 79.08 & NR & -- & -- & --\\
 & 3 & 68.25 & NR & NR & -- & --\\
 & 4 & 59.08 & 62.58 & 60.08 & 86.17 & --\\
 & 5 & 48.08 & 58.50 & 54.33 & 57.50 & 82.42\\
\addlinespace
\end{longtable}
\endgroup

\begingroup\small\setlength{\tabcolsep}{4pt}\setlength{\LTpre}{6pt}\setlength{\LTpost}{6pt}
\begin{longtable}{llrrrr}
\caption{CIFAR-100 baseline task-accuracy matrix. Rows indicate the task just completed; columns indicate the evaluated task.}\label{tab:accuracy_base100}\\
\toprule
Model & After task & T1 & T2 & T3 & T4\\\midrule
\endfirsthead
\multicolumn{6}{l}{\tablename\ \thetable\ (continued)}\\
\toprule
Model & After task & T1 & T2 & T3 & T4\\\midrule
\endhead
\midrule\multicolumn{6}{r}{Continued on next page}\\\endfoot
\bottomrule\endlastfoot
ResNet-18 & 1 & 92.62 & -- & -- & --\\
 & 2 & 76.88 & 91.00 & -- & --\\
 & 3 & 57.98 & 68.38 & 92.10 & --\\
 & 4 & 50.33 & 58.00 & 66.03 & 93.13\\
\addlinespace
DeiT-Tiny & 1 & 91.62 & -- & -- & --\\
 & 2 & 71.43 & NR & -- & --\\
 & 3 & 53.62 & 58.55 & 88.03 & --\\
 & 4 & 43.63 & 44.08 & 53.13 & 89.48\\
\addlinespace
MobileNetV3-S & 1 & 80.48 & -- & -- & --\\
 & 2 & 58.70 & NR & -- & --\\
 & 3 & 43.98 & 49.40 & 82.12 & --\\
 & 4 & 38.97 & 42.85 & 45.30 & 84.17\\
\addlinespace
MLP-Mixer-B/16 & 1 & 89.47 & -- & -- & --\\
 & 2 & 61.32 & NR & -- & --\\
 & 3 & 41.62 & 47.13 & 85.20 & --\\
 & 4 & 29.72 & 34.25 & 46.18 & 87.32\\
\addlinespace
\end{longtable}
\endgroup

\subsection{Matched early-content histories}
Each entry averages the six test histories in a branch. T1 is the shared task $A$, T2 is $B_X$ or $B_Y$, and later columns are the common tasks. The alternative second tasks have disjoint classes.
\begingroup\small\setlength{\tabcolsep}{4pt}\setlength{\LTpre}{6pt}\setlength{\LTpost}{6pt}
\begin{longtable}{lllrrrr}
\caption{CIFAR-10 paired early-content task accuracies.}\label{tab:accuracy_content10}\\
\toprule
Model & Branch & After task & T1 & T2 & T3 & T4\\\midrule
\endfirsthead
\multicolumn{7}{l}{\tablename\ \thetable\ (continued)}\\
\toprule
Model & Branch & After task & T1 & T2 & T3 & T4\\\midrule
\endhead
\midrule\multicolumn{7}{r}{Continued on next page}\\\endfoot
\bottomrule\endlastfoot
ResNet-18 & X & 1 & 95.17 & -- & -- & --\\
 &  & 2 & 96.83 & 96.83 & -- & --\\
 &  & 3 & 91.50 & 0.00 & 95.50 & --\\
 &  & 4 & 85.83 & 0.00 & 90.50 & 79.33\\
 & Y & 1 & 95.17 & -- & -- & --\\
 &  & 2 & 92.00 & 96.33 & -- & --\\
 &  & 3 & 90.83 & 0.00 & 94.67 & --\\
 &  & 4 & 84.83 & 0.00 & 89.00 & 80.83\\
\addlinespace
DeiT-Tiny & X & 1 & 94.00 & -- & -- & --\\
 &  & 2 & 92.67 & 96.00 & -- & --\\
 &  & 3 & 90.33 & 0.00 & 92.67 & --\\
 &  & 4 & 73.67 & 0.00 & 81.67 & 76.83\\
 & Y & 1 & 94.00 & -- & -- & --\\
 &  & 2 & 89.50 & 94.00 & -- & --\\
 &  & 3 & 87.00 & 0.00 & 91.50 & --\\
 &  & 4 & 79.83 & 0.00 & 82.00 & 76.67\\
\addlinespace
MobileNetV3-S & X & 1 & 86.83 & -- & -- & --\\
 &  & 2 & 86.33 & 94.17 & -- & --\\
 &  & 3 & 77.00 & 0.00 & 90.17 & --\\
 &  & 4 & 74.00 & 0.00 & 81.83 & 68.33\\
 & Y & 1 & 86.83 & -- & -- & --\\
 &  & 2 & 80.50 & 91.67 & -- & --\\
 &  & 3 & 79.00 & 0.00 & 77.33 & --\\
 &  & 4 & 69.83 & 0.00 & 78.33 & 73.00\\
\addlinespace
MLP-Mixer-B/16 & X & 1 & 80.50 & -- & -- & --\\
 &  & 2 & 85.83 & 91.17 & -- & --\\
 &  & 3 & 80.67 & 0.00 & 88.00 & --\\
 &  & 4 & 63.50 & 0.00 & 73.83 & 66.50\\
 & Y & 1 & 80.50 & -- & -- & --\\
 &  & 2 & 80.00 & 77.67 & -- & --\\
 &  & 3 & 74.83 & 0.00 & 83.67 & --\\
 &  & 4 & 65.17 & 0.00 & 72.67 & 64.67\\
\addlinespace
\end{longtable}
\endgroup

\begingroup\small\setlength{\tabcolsep}{4pt}\setlength{\LTpre}{6pt}\setlength{\LTpost}{6pt}
\begin{longtable}{lllrrr}
\caption{CIFAR-100 paired early-content task accuracies.}\label{tab:accuracy_content100}\\
\toprule
Model & Branch & After task & T1 & T2 & T3\\\midrule
\endfirsthead
\multicolumn{6}{l}{\tablename\ \thetable\ (continued)}\\
\toprule
Model & Branch & After task & T1 & T2 & T3\\\midrule
\endhead
\midrule\multicolumn{6}{r}{Continued on next page}\\\endfoot
\bottomrule\endlastfoot
ResNet-18 & X & 1 & 92.20 & -- & --\\
 &  & 2 & 75.77 & 94.37 & --\\
 &  & 3 & 67.77 & 4.77 & 89.07\\
 & Y & 1 & 92.20 & -- & --\\
 &  & 2 & 77.87 & 92.07 & --\\
 &  & 3 & 68.10 & 5.97 & 88.90\\
\addlinespace
DeiT-Tiny & X & 1 & 91.40 & -- & --\\
 &  & 2 & 69.43 & 89.47 & --\\
 &  & 3 & 62.97 & 3.57 & 84.43\\
 & Y & 1 & 91.40 & -- & --\\
 &  & 2 & 71.93 & 89.83 & --\\
 &  & 3 & 65.50 & 3.07 & 83.50\\
\addlinespace
MobileNetV3-S & X & 1 & 82.90 & -- & --\\
 &  & 2 & 57.40 & 82.17 & --\\
 &  & 3 & 52.40 & 0.40 & 75.83\\
 & Y & 1 & 82.90 & -- & --\\
 &  & 2 & 66.17 & 77.87 & --\\
 &  & 3 & 58.23 & 1.47 & 78.33\\
\addlinespace
MLP-Mixer-B/16 & X & 1 & 88.67 & -- & --\\
 &  & 2 & 62.87 & 87.67 & --\\
 &  & 3 & 54.57 & 2.80 & 79.87\\
 & Y & 1 & 88.67 & -- & --\\
 &  & 2 & 64.00 & 88.03 & --\\
 &  & 3 & 55.70 & 4.67 & 80.20\\
\addlinespace
\end{longtable}
\endgroup

\subsection{Learning without replay}
\begingroup\small\setlength{\tabcolsep}{4pt}\setlength{\LTpre}{6pt}\setlength{\LTpost}{6pt}
\begin{longtable}{llrrrrr}
\caption{CIFAR-10 task accuracies without replay, averaged over twelve test histories.}\label{tab:accuracy_noreplay}\\
\toprule
Model & After task & T1 & T2 & T3 & T4 & T5\\\midrule
\endfirsthead
\multicolumn{7}{l}{\tablename\ \thetable\ (continued)}\\
\toprule
Model & After task & T1 & T2 & T3 & T4 & T5\\\midrule
\endhead
\midrule\multicolumn{7}{r}{Continued on next page}\\\endfoot
\bottomrule\endlastfoot
ResNet-18 & 1 & 93.00 & -- & -- & -- & --\\
 & 2 & 0.00 & 96.75 & -- & -- & --\\
 & 3 & 0.00 & 0.00 & 96.83 & -- & --\\
 & 4 & 0.00 & 0.00 & 0.00 & 96.17 & --\\
 & 5 & 0.00 & 0.00 & 0.00 & 0.00 & 92.08\\
\addlinespace
DeiT-Tiny & 1 & 85.17 & -- & -- & -- & --\\
 & 2 & 0.00 & 95.08 & -- & -- & --\\
 & 3 & 0.00 & 0.00 & 91.83 & -- & --\\
 & 4 & 0.00 & 0.00 & 0.00 & 90.08 & --\\
 & 5 & 0.00 & 0.00 & 0.00 & 0.00 & 84.00\\
\addlinespace
MobileNetV3-S & 1 & 86.58 & -- & -- & -- & --\\
 & 2 & 0.00 & 87.75 & -- & -- & --\\
 & 3 & 0.00 & 0.00 & 89.83 & -- & --\\
 & 4 & 0.00 & 0.00 & 0.00 & 88.42 & --\\
 & 5 & 0.00 & 0.00 & 0.00 & 0.00 & 83.58\\
\addlinespace
MLP-Mixer-B/16 & 1 & 83.00 & -- & -- & -- & --\\
 & 2 & 0.00 & 81.33 & -- & -- & --\\
 & 3 & 0.00 & 0.00 & 79.92 & -- & --\\
 & 4 & 0.00 & 0.00 & 0.00 & 81.58 & --\\
 & 5 & 0.00 & 0.00 & 0.00 & 0.00 & 75.58\\
\addlinespace
\end{longtable}
\endgroup

\subsection{Absolute accuracies in the displayed intervention}
These summaries use coefficient one hundred, matching Table~\ref{tab:intervention}. New, Old and All average the final task, all preceding tasks, and all seen tasks, respectively. Mean denotes the average over epochs one through five; Final denotes the fifth epoch. Random is the within-history average of the two random constraints. All entries then average six histories.
\begingroup\small\setlength{\tabcolsep}{4pt}\setlength{\LTpre}{6pt}\setlength{\LTpost}{6pt}
\begin{longtable}{llrrrrrr}
\caption{CIFAR-10 absolute intervention accuracies at $\lambda=100$. First denotes initial-task retention.}\label{tab:accuracy_primary10}\\
\toprule
Model & Branch & Mean New & Mean Old & Mean All & Final First & Final Old & Final All\\\midrule
\endfirsthead
\multicolumn{8}{l}{\tablename\ \thetable\ (continued)}\\
\toprule
Model & Branch & Mean New & Mean Old & Mean All & Final First & Final Old & Final All\\\midrule
\endhead
\midrule\multicolumn{8}{r}{Continued on next page}\\\endfoot
\bottomrule\endlastfoot
ResNet-18 & Normal & 88.37 & 71.63 & 74.98 & 63.00 & 74.08 & 77.63\\
 & Scaffold & 67.43 & 70.12 & 69.58 & 63.33 & 68.83 & 70.10\\
 & Random 1 & 74.47 & 70.07 & 70.95 & 61.83 & 69.88 & 72.60\\
 & Random 2 & 74.33 & 69.78 & 70.69 & 61.50 & 68.58 & 71.37\\
 & Random & 74.40 & 69.93 & 70.82 & 61.67 & 69.23 & 71.98\\
\addlinespace
DeiT-Tiny & Normal & 89.63 & 70.71 & 74.49 & 57.33 & 67.92 & 72.70\\
 & Scaffold & 70.60 & 69.66 & 69.85 & 59.00 & 68.96 & 71.63\\
 & Random 1 & 71.13 & 70.26 & 70.43 & 57.17 & 69.04 & 71.70\\
 & Random 2 & 70.77 & 70.02 & 70.17 & 57.83 & 69.58 & 72.10\\
 & Random & 70.95 & 70.14 & 70.30 & 57.50 & 69.31 & 71.90\\
\addlinespace
MobileNetV3-S & Normal & 84.50 & 60.44 & 65.25 & 50.33 & 58.17 & 63.03\\
 & Scaffold & 76.13 & 57.31 & 61.07 & 49.17 & 58.33 & 63.17\\
 & Random 1 & 79.80 & 58.07 & 62.41 & 46.17 & 58.42 & 64.17\\
 & Random 2 & 81.97 & 58.28 & 63.02 & 46.83 & 59.67 & 65.07\\
 & Random & 80.88 & 58.18 & 62.72 & 46.50 & 59.04 & 64.62\\
\addlinespace
MLP-Mixer-B/16 & Normal & 81.43 & 55.67 & 60.82 & 45.50 & 56.92 & 61.93\\
 & Scaffold & 64.17 & 55.02 & 56.85 & 47.50 & 57.58 & 59.97\\
 & Random 1 & 72.30 & 50.85 & 55.14 & 46.17 & 55.58 & 59.40\\
 & Random 2 & 72.67 & 50.39 & 54.85 & 46.67 & 56.00 & 59.80\\
 & Random & 72.48 & 50.62 & 54.99 & 46.42 & 55.79 & 59.60\\
\addlinespace
\end{longtable}
\endgroup

\begingroup\small\setlength{\tabcolsep}{4pt}\setlength{\LTpre}{6pt}\setlength{\LTpost}{6pt}
\begin{longtable}{llrrrrrr}
\caption{CIFAR-100 absolute intervention accuracies at $\lambda=100$. First denotes initial-task retention.}\label{tab:accuracy_primary100}\\
\toprule
Model & Branch & Mean New & Mean Old & Mean All & Final First & Final Old & Final All\\\midrule
\endfirsthead
\multicolumn{8}{l}{\tablename\ \thetable\ (continued)}\\
\toprule
Model & Branch & Mean New & Mean Old & Mean All & Final First & Final Old & Final All\\\midrule
\endhead
\midrule\multicolumn{8}{r}{Continued on next page}\\\endfoot
\bottomrule\endlastfoot
ResNet-18 & Normal & 91.55 & 58.66 & 66.88 & 51.63 & 58.33 & 67.09\\
 & Scaffold & 77.44 & 61.65 & 65.60 & 49.87 & 61.41 & 66.73\\
 & Random 1 & 83.75 & 61.64 & 67.17 & 49.87 & 61.38 & 68.31\\
 & Random 2 & 83.27 & 61.40 & 66.87 & 49.53 & 60.82 & 67.70\\
 & Random & 83.51 & 61.52 & 67.02 & 49.70 & 61.10 & 68.00\\
\addlinespace
DeiT-Tiny & Normal & 89.55 & 49.67 & 59.64 & 41.20 & 46.46 & 57.09\\
 & Scaffold & 82.81 & 57.22 & 63.61 & 44.20 & 56.83 & 64.47\\
 & Random 1 & 83.43 & 56.82 & 63.47 & 44.60 & 56.57 & 64.44\\
 & Random 2 & 83.77 & 56.84 & 63.57 & 44.07 & 56.49 & 64.34\\
 & Random & 83.60 & 56.83 & 63.52 & 44.33 & 56.53 & 64.39\\
\addlinespace
MobileNetV3-S & Normal & 84.88 & 43.21 & 53.63 & 36.30 & 41.22 & 52.29\\
 & Scaffold & 74.17 & 50.79 & 56.64 & 42.57 & 50.34 & 57.64\\
 & Random 1 & 75.48 & 49.66 & 56.12 & 41.30 & 50.57 & 57.78\\
 & Random 2 & 74.61 & 49.62 & 55.86 & 40.13 & 49.74 & 56.72\\
 & Random & 75.04 & 49.64 & 55.99 & 40.72 & 50.16 & 57.25\\
\addlinespace
MLP-Mixer-B/16 & Normal & 86.62 & 39.85 & 51.54 & 31.13 & 38.42 & 50.69\\
 & Scaffold & 81.01 & 49.18 & 57.13 & 34.17 & 48.53 & 57.67\\
 & Random 1 & 84.63 & 46.01 & 55.67 & 30.33 & 44.02 & 55.43\\
 & Random 2 & 84.36 & 46.22 & 55.75 & 31.40 & 45.27 & 56.40\\
 & Random & 84.50 & 46.11 & 55.71 & 30.87 & 44.64 & 55.91\\
\addlinespace
\end{longtable}
\endgroup

\Needspace{12\baselineskip}
\begingroup\small\setlength{\tabcolsep}{4pt}\setlength{\LTpre}{6pt}\setlength{\LTpost}{6pt}
\begin{longtable}{llrrr}
\caption{Final old-task retention: scaffold minus random (R) or normal training (B), in percentage points [95\% paired hierarchical interval], at coefficient one hundred. Old averages every preceding task.}\label{tab:old_retention}\\
\toprule
Model & CIFAR & First versus R & Old versus R & Old versus B\\\midrule
\endfirsthead
\multicolumn{5}{l}{\tablename\ \thetable\ (continued)}\\
\toprule
Model & CIFAR & First versus R & Old versus R & Old versus B\\\midrule
\endhead
\midrule\multicolumn{5}{r}{Continued on next page}\\\endfoot
\bottomrule\endlastfoot
ResNet-18 & 10 & +1.67 [-1.00, +4.83] & -0.40 [-1.75, +0.62] & -5.25 [-8.75, -1.37]\\
ResNet-18 & 100 & +0.17 [-0.85, +1.23] & +0.31 [-0.47, +1.08] & +3.08 [+1.90, +4.33]\\
DeiT-Tiny & 10 & +1.50 [-2.50, +4.50] & -0.35 [-1.58, +0.48] & +1.04 [-2.67, +4.54]\\
DeiT-Tiny & 100 & -0.13 [-1.10, +0.77] & +0.31 [-0.46, +1.10] & +10.38 [+8.13, +13.39]\\
MobileNetV3-S & 10 & +2.67 [-1.17, +6.00] & -0.71 [-3.92, +2.13] & +0.17 [-5.33, +5.21]\\
MobileNetV3-S & 100 & +1.85 [+0.00, +4.27] & +0.19 [-1.90, +2.50] & +9.12 [+3.47, +14.04]\\
MLP-Mixer-B/16 & 10 & +1.08 [-6.00, +6.17] & +1.79 [-1.77, +5.54] & +0.67 [-5.50, +5.25]\\
MLP-Mixer-B/16 & 100 & +3.30 [+1.40, +5.45] & +3.89 [+1.83, +6.23] & +10.11 [+7.78, +12.41]\\
\end{longtable}
\endgroup

\subsection{Every task across the intervention epochs}
The following tables report the fixed-coefficient interventions at $\lambda=10$ and 100, with normal training as the reference. Normal denotes unconstrained training; Scaffold, R1 and R2 denote the scaffold and two random constraints. Each entry averages six paired histories. Mean averages the five displayed epochs, excluding the pre-task measurement. Old averages all preceding tasks; All averages all tasks. The last task column measures new-task learning.
\begingroup\small\setlength{\tabcolsep}{4pt}\setlength{\LTpre}{6pt}\setlength{\LTpost}{6pt}
\begin{longtable}{lllrrrrrrr}
\caption{ResNet-18, CIFAR-10: task accuracies during the final-task intervention at coefficients 10 and 100.}\label{tab:accuracy_epochs_resnet18_cifar10}\\
\toprule
Branch & $\lambda$ & Epoch & T1 & T2 & T3 & T4 & T5 & Old & All\\\midrule
\endfirsthead
\multicolumn{10}{l}{\tablename\ \thetable\ (continued)}\\
\toprule
Branch & $\lambda$ & Epoch & T1 & T2 & T3 & T4 & T5 & Old & All\\\midrule
\endhead
\midrule\multicolumn{10}{r}{Continued on next page}\\\endfoot
\bottomrule\endlastfoot
Normal & 0 & 1 & 60.50 & 65.83 & 64.67 & 71.17 & 81.50 & 65.54 & 68.73\\
Normal & 0 & 2 & 62.67 & 71.00 & 72.83 & 84.50 & 86.33 & 72.75 & 75.47\\
Normal & 0 & 3 & 62.00 & 69.67 & 72.67 & 83.83 & 89.67 & 72.04 & 75.57\\
Normal & 0 & 4 & 64.50 & 71.00 & 72.67 & 86.83 & 92.50 & 73.75 & 77.50\\
Normal & 0 & 5 & 63.00 & 73.67 & 75.00 & 84.67 & 91.83 & 74.08 & 77.63\\
Normal & 0 & Mean & 62.53 & 70.23 & 71.57 & 82.20 & 88.37 & 71.63 & 74.98\\
\addlinespace
Scaffold & 10 & 1 & 68.00 & 70.67 & 72.50 & 80.67 & 61.00 & 72.96 & 70.57\\
Scaffold & 10 & 2 & 66.67 & 66.67 & 72.67 & 71.83 & 72.17 & 69.46 & 70.00\\
Scaffold & 10 & 3 & 64.33 & 67.67 & 72.67 & 73.00 & 78.50 & 69.42 & 71.23\\
Scaffold & 10 & 4 & 61.67 & 69.33 & 74.33 & 74.67 & 80.00 & 70.00 & 72.00\\
Scaffold & 10 & 5 & 63.83 & 69.67 & 73.33 & 76.33 & 82.00 & 70.79 & 73.03\\
Scaffold & 10 & Mean & 64.90 & 68.80 & 73.10 & 75.30 & 74.73 & 70.52 & 71.37\\
\addlinespace
R1 & 10 & 1 & 64.00 & 67.17 & 70.50 & 75.33 & 79.00 & 69.25 & 71.20\\
R1 & 10 & 2 & 63.50 & 68.67 & 74.17 & 74.67 & 84.00 & 70.25 & 73.00\\
R1 & 10 & 3 & 63.00 & 68.83 & 71.50 & 75.83 & 89.33 & 69.79 & 73.70\\
R1 & 10 & 4 & 63.33 & 68.33 & 72.00 & 80.33 & 90.67 & 71.00 & 74.93\\
R1 & 10 & 5 & 63.00 & 70.67 & 73.17 & 81.67 & 92.00 & 72.13 & 76.10\\
R1 & 10 & Mean & 63.37 & 68.73 & 72.27 & 77.57 & 87.00 & 70.48 & 73.79\\
\addlinespace
R2 & 10 & 1 & 65.50 & 66.67 & 70.83 & 74.00 & 77.17 & 69.25 & 70.83\\
R2 & 10 & 2 & 63.33 & 69.83 & 71.83 & 73.67 & 83.83 & 69.67 & 72.50\\
R2 & 10 & 3 & 64.50 & 68.83 & 71.50 & 77.17 & 88.33 & 70.50 & 74.07\\
R2 & 10 & 4 & 64.33 & 68.50 & 70.17 & 80.00 & 90.00 & 70.75 & 74.60\\
R2 & 10 & 5 & 63.33 & 70.33 & 72.50 & 80.83 & 90.83 & 71.75 & 75.57\\
R2 & 10 & Mean & 64.20 & 68.83 & 71.37 & 77.13 & 86.03 & 70.38 & 73.51\\
\addlinespace
Scaffold & 100 & 1 & 68.00 & 72.17 & 74.67 & 85.17 & 47.50 & 75.00 & 69.50\\
Scaffold & 100 & 2 & 65.33 & 68.33 & 71.83 & 71.33 & 68.33 & 69.21 & 69.03\\
Scaffold & 100 & 3 & 62.83 & 67.33 & 71.50 & 72.83 & 73.00 & 68.62 & 69.50\\
Scaffold & 100 & 4 & 65.17 & 67.67 & 70.17 & 72.67 & 73.17 & 68.92 & 69.77\\
Scaffold & 100 & 5 & 63.33 & 67.33 & 70.00 & 74.67 & 75.17 & 68.83 & 70.10\\
Scaffold & 100 & Mean & 64.93 & 68.57 & 71.63 & 75.33 & 67.43 & 70.12 & 69.58\\
\addlinespace
R1 & 100 & 1 & 65.50 & 69.50 & 73.67 & 82.00 & 56.17 & 72.67 & 69.37\\
R1 & 100 & 2 & 65.67 & 67.50 & 72.33 & 72.83 & 74.00 & 69.58 & 70.47\\
R1 & 100 & 3 & 65.33 & 66.83 & 71.17 & 73.83 & 77.33 & 69.29 & 70.90\\
R1 & 100 & 4 & 62.83 & 66.33 & 71.83 & 74.67 & 81.33 & 68.92 & 71.40\\
R1 & 100 & 5 & 61.83 & 67.50 & 71.33 & 78.83 & 83.50 & 69.88 & 72.60\\
R1 & 100 & Mean & 64.23 & 67.53 & 72.07 & 76.43 & 74.47 & 70.07 & 70.95\\
\addlinespace
R2 & 100 & 1 & 67.17 & 70.83 & 72.17 & 83.33 & 58.33 & 73.37 & 70.37\\
R2 & 100 & 2 & 65.00 & 67.00 & 71.33 & 72.50 & 74.50 & 68.96 & 70.07\\
R2 & 100 & 3 & 64.83 & 66.67 & 72.33 & 72.00 & 78.00 & 68.96 & 70.77\\
R2 & 100 & 4 & 61.83 & 67.67 & 71.17 & 75.50 & 78.33 & 69.04 & 70.90\\
R2 & 100 & 5 & 61.50 & 67.17 & 70.50 & 75.17 & 82.50 & 68.58 & 71.37\\
R2 & 100 & Mean & 64.07 & 67.87 & 71.50 & 75.70 & 74.33 & 69.78 & 70.69\\
\addlinespace
\end{longtable}
\endgroup

\begingroup\small\setlength{\tabcolsep}{4pt}\setlength{\LTpre}{6pt}\setlength{\LTpost}{6pt}
\begin{longtable}{lllrrrrrr}
\caption{ResNet-18, CIFAR-100: task accuracies during the final-task intervention at coefficients 10 and 100.}\label{tab:accuracy_epochs_resnet18_cifar100}\\
\toprule
Branch & $\lambda$ & Epoch & T1 & T2 & T3 & T4 & Old & All\\\midrule
\endfirsthead
\multicolumn{9}{l}{\tablename\ \thetable\ (continued)}\\
\toprule
Branch & $\lambda$ & Epoch & T1 & T2 & T3 & T4 & Old & All\\\midrule
\endhead
\midrule\multicolumn{9}{r}{Continued on next page}\\\endfoot
\bottomrule\endlastfoot
Normal & 0 & 1 & 50.03 & 54.57 & 75.90 & 88.00 & 60.17 & 67.12\\
Normal & 0 & 2 & 49.60 & 53.60 & 72.93 & 91.57 & 58.71 & 66.92\\
Normal & 0 & 3 & 49.83 & 54.67 & 70.90 & 92.40 & 58.47 & 66.95\\
Normal & 0 & 4 & 50.10 & 53.83 & 68.97 & 92.40 & 57.63 & 66.32\\
Normal & 0 & 5 & 51.63 & 55.20 & 68.17 & 93.37 & 58.33 & 67.09\\
Normal & 0 & Mean & 50.24 & 54.37 & 71.37 & 91.55 & 58.66 & 66.88\\
\addlinespace
Scaffold & 10 & 1 & 50.93 & 56.97 & 79.97 & 73.17 & 62.62 & 65.26\\
Scaffold & 10 & 2 & 49.33 & 55.13 & 78.20 & 84.00 & 60.89 & 66.67\\
Scaffold & 10 & 3 & 48.53 & 55.10 & 78.20 & 86.40 & 60.61 & 67.06\\
Scaffold & 10 & 4 & 50.20 & 56.10 & 77.80 & 88.40 & 61.37 & 68.13\\
Scaffold & 10 & 5 & 51.23 & 55.90 & 77.63 & 88.90 & 61.59 & 68.42\\
Scaffold & 10 & Mean & 50.05 & 55.84 & 78.36 & 84.17 & 61.42 & 67.11\\
\addlinespace
R1 & 10 & 1 & 49.43 & 55.17 & 77.43 & 86.77 & 60.68 & 67.20\\
R1 & 10 & 2 & 50.17 & 54.53 & 75.03 & 90.93 & 59.91 & 67.67\\
R1 & 10 & 3 & 50.60 & 55.40 & 74.80 & 92.27 & 60.27 & 68.27\\
R1 & 10 & 4 & 51.40 & 57.00 & 74.30 & 92.97 & 60.90 & 68.92\\
R1 & 10 & 5 & 51.73 & 57.07 & 73.20 & 92.87 & 60.67 & 68.72\\
R1 & 10 & Mean & 50.67 & 55.83 & 74.95 & 91.16 & 60.48 & 68.15\\
\addlinespace
R2 & 10 & 1 & 50.63 & 55.37 & 77.63 & 85.70 & 61.21 & 67.33\\
R2 & 10 & 2 & 50.70 & 54.17 & 75.53 & 90.63 & 60.13 & 67.76\\
R2 & 10 & 3 & 50.83 & 55.50 & 74.10 & 91.80 & 60.14 & 68.06\\
R2 & 10 & 4 & 51.33 & 56.47 & 73.70 & 92.47 & 60.50 & 68.49\\
R2 & 10 & 5 & 51.47 & 55.80 & 73.33 & 92.77 & 60.20 & 68.34\\
R2 & 10 & Mean & 50.99 & 55.46 & 74.86 & 90.67 & 60.44 & 68.00\\
\addlinespace
Scaffold & 100 & 1 & 51.70 & 57.87 & 80.87 & 66.03 & 63.48 & 64.12\\
Scaffold & 100 & 2 & 49.10 & 55.10 & 78.37 & 76.93 & 60.86 & 64.88\\
Scaffold & 100 & 3 & 49.53 & 55.27 & 78.07 & 79.90 & 60.96 & 65.69\\
Scaffold & 100 & 4 & 49.63 & 56.67 & 78.33 & 81.63 & 61.54 & 66.57\\
Scaffold & 100 & 5 & 49.87 & 56.33 & 78.03 & 82.70 & 61.41 & 66.73\\
Scaffold & 100 & Mean & 49.97 & 56.25 & 78.73 & 77.44 & 61.65 & 65.60\\
\addlinespace
R1 & 100 & 1 & 52.57 & 58.27 & 81.03 & 71.20 & 63.96 & 65.77\\
R1 & 100 & 2 & 49.40 & 54.63 & 78.33 & 83.47 & 60.79 & 66.46\\
R1 & 100 & 3 & 48.77 & 55.30 & 78.43 & 87.17 & 60.83 & 67.42\\
R1 & 100 & 4 & 49.97 & 56.43 & 77.40 & 87.83 & 61.27 & 67.91\\
R1 & 100 & 5 & 49.87 & 56.60 & 77.67 & 89.10 & 61.38 & 68.31\\
R1 & 100 & Mean & 50.11 & 56.25 & 78.57 & 83.75 & 61.64 & 67.17\\
\addlinespace
R2 & 100 & 1 & 52.17 & 57.60 & 80.57 & 71.57 & 63.44 & 65.48\\
R2 & 100 & 2 & 49.27 & 54.13 & 78.80 & 82.97 & 60.73 & 66.29\\
R2 & 100 & 3 & 49.10 & 55.17 & 78.27 & 85.87 & 60.84 & 67.10\\
R2 & 100 & 4 & 49.67 & 56.23 & 77.53 & 87.63 & 61.14 & 67.77\\
R2 & 100 & 5 & 49.53 & 55.70 & 77.23 & 88.33 & 60.82 & 67.70\\
R2 & 100 & Mean & 49.95 & 55.77 & 78.48 & 83.27 & 61.40 & 66.87\\
\addlinespace
\end{longtable}
\endgroup

\begingroup\small\setlength{\tabcolsep}{4pt}\setlength{\LTpre}{6pt}\setlength{\LTpost}{6pt}
\begin{longtable}{lllrrrrrrr}
\caption{DeiT-Tiny, CIFAR-10: task accuracies during the final-task intervention at coefficients 10 and 100.}\label{tab:accuracy_epochs_deit_tiny_cifar10}\\
\toprule
Branch & $\lambda$ & Epoch & T1 & T2 & T3 & T4 & T5 & Old & All\\\midrule
\endfirsthead
\multicolumn{10}{l}{\tablename\ \thetable\ (continued)}\\
\toprule
Branch & $\lambda$ & Epoch & T1 & T2 & T3 & T4 & T5 & Old & All\\\midrule
\endhead
\midrule\multicolumn{10}{r}{Continued on next page}\\\endfoot
\bottomrule\endlastfoot
Normal & 0 & 1 & 61.50 & 69.00 & 72.67 & 77.83 & 84.17 & 70.25 & 73.03\\
Normal & 0 & 2 & 60.67 & 71.33 & 72.83 & 83.00 & 91.00 & 71.96 & 75.77\\
Normal & 0 & 3 & 61.33 & 73.67 & 72.33 & 82.67 & 91.33 & 72.50 & 76.27\\
Normal & 0 & 4 & 57.50 & 70.33 & 74.67 & 81.17 & 89.83 & 70.92 & 74.70\\
Normal & 0 & 5 & 57.33 & 67.17 & 68.33 & 78.83 & 91.83 & 67.92 & 72.70\\
Normal & 0 & Mean & 59.67 & 70.30 & 72.17 & 80.70 & 89.63 & 70.71 & 74.49\\
\addlinespace
Scaffold & 10 & 1 & 55.33 & 59.50 & 67.50 & 75.33 & 79.83 & 64.42 & 67.50\\
Scaffold & 10 & 2 & 63.17 & 66.50 & 72.83 & 82.67 & 84.17 & 71.29 & 73.87\\
Scaffold & 10 & 3 & 58.00 & 69.00 & 74.00 & 83.00 & 87.67 & 71.00 & 74.33\\
Scaffold & 10 & 4 & 59.00 & 70.83 & 74.17 & 84.83 & 89.67 & 72.21 & 75.70\\
Scaffold & 10 & 5 & 61.33 & 70.67 & 74.33 & 86.83 & 91.67 & 73.29 & 76.97\\
Scaffold & 10 & Mean & 59.37 & 67.30 & 72.57 & 82.53 & 86.60 & 70.44 & 73.67\\
\addlinespace
R1 & 10 & 1 & 58.17 & 68.17 & 68.00 & 76.67 & 79.83 & 67.75 & 70.17\\
R1 & 10 & 2 & 53.83 & 69.67 & 70.00 & 82.83 & 85.17 & 69.08 & 72.30\\
R1 & 10 & 3 & 61.00 & 70.17 & 74.67 & 82.50 & 89.50 & 72.08 & 75.57\\
R1 & 10 & 4 & 59.83 & 69.00 & 74.33 & 83.67 & 90.83 & 71.71 & 75.53\\
R1 & 10 & 5 & 60.17 & 70.83 & 75.50 & 84.17 & 91.67 & 72.67 & 76.47\\
R1 & 10 & Mean & 58.60 & 69.57 & 72.50 & 81.97 & 87.40 & 70.66 & 74.01\\
\addlinespace
R2 & 10 & 1 & 59.33 & 61.67 & 66.17 & 77.67 & 78.17 & 66.21 & 68.60\\
R2 & 10 & 2 & 60.83 & 67.00 & 70.67 & 82.67 & 86.00 & 70.29 & 73.43\\
R2 & 10 & 3 & 58.00 & 69.33 & 73.83 & 85.67 & 89.50 & 71.71 & 75.27\\
R2 & 10 & 4 & 60.67 & 69.83 & 73.83 & 85.83 & 90.33 & 72.54 & 76.10\\
R2 & 10 & 5 & 60.83 & 72.33 & 75.33 & 86.83 & 91.17 & 73.83 & 77.30\\
R2 & 10 & Mean & 59.93 & 68.03 & 71.97 & 83.73 & 87.03 & 70.92 & 74.14\\
\addlinespace
Scaffold & 100 & 1 & 67.67 & 72.83 & 79.00 & 93.33 & 36.33 & 78.21 & 69.83\\
Scaffold & 100 & 2 & 63.50 & 65.17 & 63.50 & 76.83 & 76.00 & 67.25 & 69.00\\
Scaffold & 100 & 3 & 55.83 & 63.33 & 66.67 & 77.67 & 78.17 & 65.88 & 68.33\\
Scaffold & 100 & 4 & 57.50 & 65.17 & 67.33 & 82.00 & 80.17 & 68.00 & 70.43\\
Scaffold & 100 & 5 & 59.00 & 65.83 & 69.17 & 81.83 & 82.33 & 68.96 & 71.63\\
Scaffold & 100 & Mean & 60.70 & 66.47 & 69.13 & 82.33 & 70.60 & 69.66 & 69.85\\
\addlinespace
R1 & 100 & 1 & 68.33 & 78.67 & 81.67 & 90.00 & 37.50 & 79.67 & 71.23\\
R1 & 100 & 2 & 57.50 & 64.17 & 65.17 & 77.50 & 76.17 & 66.08 & 68.10\\
R1 & 100 & 3 & 57.00 & 66.17 & 67.83 & 78.50 & 79.50 & 67.38 & 69.80\\
R1 & 100 & 4 & 58.67 & 65.83 & 70.83 & 81.17 & 80.17 & 69.13 & 71.33\\
R1 & 100 & 5 & 57.17 & 67.67 & 71.33 & 80.00 & 82.33 & 69.04 & 71.70\\
R1 & 100 & Mean & 59.73 & 68.50 & 71.37 & 81.43 & 71.13 & 70.26 & 70.43\\
\addlinespace
R2 & 100 & 1 & 68.83 & 74.33 & 81.17 & 91.00 & 36.33 & 78.83 & 70.33\\
R2 & 100 & 2 & 58.17 & 64.33 & 64.33 & 75.50 & 74.83 & 65.58 & 67.43\\
R2 & 100 & 3 & 56.67 & 66.33 & 68.17 & 78.17 & 79.00 & 67.33 & 69.67\\
R2 & 100 & 4 & 56.50 & 67.17 & 70.33 & 81.00 & 81.50 & 68.75 & 71.30\\
R2 & 100 & 5 & 57.83 & 68.33 & 70.67 & 81.50 & 82.17 & 69.58 & 72.10\\
R2 & 100 & Mean & 59.60 & 68.10 & 70.93 & 81.43 & 70.77 & 70.02 & 70.17\\
\addlinespace
\end{longtable}
\endgroup

\begingroup\small\setlength{\tabcolsep}{4pt}\setlength{\LTpre}{6pt}\setlength{\LTpost}{6pt}
\begin{longtable}{lllrrrrrr}
\caption{DeiT-Tiny, CIFAR-100: task accuracies during the final-task intervention at coefficients 10 and 100.}\label{tab:accuracy_epochs_deit_tiny_cifar100}\\
\toprule
Branch & $\lambda$ & Epoch & T1 & T2 & T3 & T4 & Old & All\\\midrule
\endfirsthead
\multicolumn{9}{l}{\tablename\ \thetable\ (continued)}\\
\toprule
Branch & $\lambda$ & Epoch & T1 & T2 & T3 & T4 & Old & All\\\midrule
\endhead
\midrule\multicolumn{9}{r}{Continued on next page}\\\endfoot
\bottomrule\endlastfoot
Normal & 0 & 1 & 45.47 & 47.50 & 66.47 & 89.37 & 53.14 & 62.20\\
Normal & 0 & 2 & 44.17 & 46.80 & 64.23 & 89.33 & 51.73 & 61.13\\
Normal & 0 & 3 & 42.97 & 45.73 & 61.40 & 89.47 & 50.03 & 59.89\\
Normal & 0 & 4 & 39.50 & 44.20 & 57.23 & 90.60 & 46.98 & 57.88\\
Normal & 0 & 5 & 41.20 & 43.63 & 54.53 & 89.00 & 46.46 & 57.09\\
Normal & 0 & Mean & 42.66 & 45.57 & 60.77 & 89.55 & 49.67 & 59.64\\
\addlinespace
Scaffold & 10 & 1 & 45.47 & 50.47 & 77.60 & 85.67 & 57.84 & 64.80\\
Scaffold & 10 & 2 & 43.73 & 50.00 & 75.07 & 90.70 & 56.27 & 64.88\\
Scaffold & 10 & 3 & 45.07 & 50.00 & 74.10 & 92.10 & 56.39 & 65.32\\
Scaffold & 10 & 4 & 45.00 & 49.73 & 73.47 & 93.10 & 56.07 & 65.33\\
Scaffold & 10 & 5 & 45.63 & 49.97 & 73.40 & 93.10 & 56.33 & 65.53\\
Scaffold & 10 & Mean & 44.98 & 50.03 & 74.73 & 90.93 & 56.58 & 65.17\\
\addlinespace
R1 & 10 & 1 & 44.90 & 50.50 & 76.73 & 86.53 & 57.38 & 64.67\\
R1 & 10 & 2 & 43.13 & 49.43 & 74.10 & 91.27 & 55.56 & 64.48\\
R1 & 10 & 3 & 45.53 & 49.77 & 73.83 & 92.67 & 56.38 & 65.45\\
R1 & 10 & 4 & 45.07 & 50.37 & 72.77 & 93.27 & 56.07 & 65.37\\
R1 & 10 & 5 & 45.93 & 49.67 & 72.63 & 93.37 & 56.08 & 65.40\\
R1 & 10 & Mean & 44.91 & 49.95 & 74.01 & 91.42 & 56.29 & 65.07\\
\addlinespace
R2 & 10 & 1 & 44.20 & 50.47 & 76.07 & 86.57 & 56.91 & 64.32\\
R2 & 10 & 2 & 44.07 & 50.87 & 74.70 & 91.60 & 56.54 & 65.31\\
R2 & 10 & 3 & 45.30 & 49.33 & 72.90 & 92.47 & 55.84 & 65.00\\
R2 & 10 & 4 & 45.50 & 49.60 & 72.67 & 93.30 & 55.92 & 65.27\\
R2 & 10 & 5 & 46.00 & 50.80 & 73.20 & 92.63 & 56.67 & 65.66\\
R2 & 10 & Mean & 45.01 & 50.21 & 73.91 & 91.31 & 56.38 & 65.11\\
\addlinespace
Scaffold & 100 & 1 & 46.50 & 50.47 & 76.63 & 73.97 & 57.87 & 61.89\\
Scaffold & 100 & 2 & 45.00 & 49.73 & 76.03 & 81.90 & 56.92 & 63.17\\
Scaffold & 100 & 3 & 45.37 & 50.43 & 76.33 & 84.67 & 57.38 & 64.20\\
Scaffold & 100 & 4 & 44.43 & 50.47 & 76.33 & 86.10 & 57.08 & 64.33\\
Scaffold & 100 & 5 & 44.20 & 49.87 & 76.43 & 87.40 & 56.83 & 64.47\\
Scaffold & 100 & Mean & 45.10 & 50.19 & 76.35 & 82.81 & 57.22 & 63.61\\
\addlinespace
R1 & 100 & 1 & 46.97 & 49.33 & 75.47 & 75.00 & 57.26 & 61.69\\
R1 & 100 & 2 & 45.23 & 49.17 & 76.33 & 82.47 & 56.91 & 63.30\\
R1 & 100 & 3 & 44.53 & 49.33 & 75.43 & 84.80 & 56.43 & 63.53\\
R1 & 100 & 4 & 45.00 & 50.23 & 75.53 & 86.80 & 56.92 & 64.39\\
R1 & 100 & 5 & 44.60 & 49.97 & 75.13 & 88.07 & 56.57 & 64.44\\
R1 & 100 & Mean & 45.27 & 49.61 & 75.58 & 83.43 & 56.82 & 63.47\\
\addlinespace
R2 & 100 & 1 & 45.90 & 49.67 & 75.47 & 76.10 & 57.01 & 61.78\\
R2 & 100 & 2 & 45.13 & 48.97 & 76.33 & 82.80 & 56.81 & 63.31\\
R2 & 100 & 3 & 45.27 & 49.70 & 76.87 & 84.97 & 57.28 & 64.20\\
R2 & 100 & 4 & 44.73 & 49.40 & 75.73 & 87.07 & 56.62 & 64.23\\
R2 & 100 & 5 & 44.07 & 49.80 & 75.60 & 87.90 & 56.49 & 64.34\\
R2 & 100 & Mean & 45.02 & 49.51 & 76.00 & 83.77 & 56.84 & 63.57\\
\addlinespace
\end{longtable}
\endgroup

\begingroup\small\setlength{\tabcolsep}{4pt}\setlength{\LTpre}{6pt}\setlength{\LTpost}{6pt}
\begin{longtable}{lllrrrrrrr}
\caption{MobileNetV3-S, CIFAR-10: task accuracies during the final-task intervention at coefficients 10 and 100.}\label{tab:accuracy_epochs_mobilenetv3_small_cifar10}\\
\toprule
Branch & $\lambda$ & Epoch & T1 & T2 & T3 & T4 & T5 & Old & All\\\midrule
\endfirsthead
\multicolumn{10}{l}{\tablename\ \thetable\ (continued)}\\
\toprule
Branch & $\lambda$ & Epoch & T1 & T2 & T3 & T4 & T5 & Old & All\\\midrule
\endhead
\midrule\multicolumn{10}{r}{Continued on next page}\\\endfoot
\bottomrule\endlastfoot
Normal & 0 & 1 & 44.83 & 55.33 & 59.83 & 64.33 & 80.00 & 56.08 & 60.87\\
Normal & 0 & 2 & 52.67 & 63.00 & 65.67 & 72.50 & 85.17 & 63.46 & 67.80\\
Normal & 0 & 3 & 55.50 & 59.67 & 61.17 & 74.33 & 87.17 & 62.67 & 67.57\\
Normal & 0 & 4 & 50.50 & 60.83 & 64.17 & 71.83 & 87.67 & 61.83 & 67.00\\
Normal & 0 & 5 & 50.33 & 46.33 & 64.33 & 71.67 & 82.50 & 58.17 & 63.03\\
Normal & 0 & Mean & 50.77 & 57.03 & 63.03 & 70.93 & 84.50 & 60.44 & 65.25\\
\addlinespace
Scaffold & 10 & 1 & 39.17 & 48.00 & 58.33 & 70.17 & 75.33 & 53.92 & 58.20\\
Scaffold & 10 & 2 & 47.00 & 52.00 & 65.33 & 69.83 & 84.50 & 58.54 & 63.73\\
Scaffold & 10 & 3 & 48.50 & 52.00 & 64.00 & 72.67 & 84.67 & 59.29 & 64.37\\
Scaffold & 10 & 4 & 49.33 & 49.83 & 62.17 & 69.67 & 84.83 & 57.75 & 63.17\\
Scaffold & 10 & 5 & 48.67 & 50.00 & 62.50 & 71.00 & 86.67 & 58.04 & 63.77\\
Scaffold & 10 & Mean & 46.53 & 50.37 & 62.47 & 70.67 & 83.20 & 57.51 & 62.65\\
\addlinespace
R1 & 10 & 1 & 42.67 & 50.00 & 65.67 & 67.00 & 79.50 & 56.33 & 60.97\\
R1 & 10 & 2 & 46.33 & 52.00 & 65.83 & 65.33 & 84.50 & 57.37 & 62.80\\
R1 & 10 & 3 & 48.67 & 49.83 & 64.00 & 68.67 & 85.83 & 57.79 & 63.40\\
R1 & 10 & 4 & 52.50 & 50.00 & 61.17 & 70.50 & 87.33 & 58.54 & 64.30\\
R1 & 10 & 5 & 48.67 & 51.50 & 63.67 & 66.83 & 85.33 & 57.67 & 63.20\\
R1 & 10 & Mean & 47.77 & 50.67 & 64.07 & 67.67 & 84.50 & 57.54 & 62.93\\
\addlinespace
R2 & 10 & 1 & 46.17 & 51.83 & 61.67 & 68.17 & 75.83 & 56.96 & 60.73\\
R2 & 10 & 2 & 49.00 & 52.17 & 66.83 & 66.00 & 83.17 & 58.50 & 63.43\\
R2 & 10 & 3 & 49.83 & 51.67 & 64.67 & 67.33 & 83.50 & 58.37 & 63.40\\
R2 & 10 & 4 & 50.83 & 50.50 & 62.83 & 66.33 & 83.50 & 57.62 & 62.80\\
R2 & 10 & 5 & 48.83 & 50.67 & 64.00 & 63.83 & 87.50 & 56.83 & 62.97\\
R2 & 10 & Mean & 48.93 & 51.37 & 64.00 & 66.33 & 82.70 & 57.66 & 62.67\\
\addlinespace
Scaffold & 100 & 1 & 37.83 & 45.00 & 52.67 & 69.83 & 65.00 & 51.33 & 54.07\\
Scaffold & 100 & 2 & 47.67 & 50.83 & 64.17 & 71.17 & 74.83 & 58.46 & 61.73\\
Scaffold & 100 & 3 & 50.17 & 52.83 & 66.83 & 71.33 & 77.33 & 60.29 & 63.70\\
Scaffold & 100 & 4 & 52.33 & 50.00 & 60.67 & 69.50 & 81.00 & 58.12 & 62.70\\
Scaffold & 100 & 5 & 49.17 & 49.83 & 64.50 & 69.83 & 82.50 & 58.33 & 63.17\\
Scaffold & 100 & Mean & 47.43 & 49.70 & 61.77 & 70.33 & 76.13 & 57.31 & 61.07\\
\addlinespace
R1 & 100 & 1 & 34.50 & 54.17 & 57.67 & 67.83 & 69.67 & 53.54 & 56.77\\
R1 & 100 & 2 & 47.17 & 52.17 & 67.50 & 68.00 & 78.50 & 58.71 & 62.67\\
R1 & 100 & 3 & 47.33 & 53.33 & 65.50 & 69.50 & 80.83 & 58.92 & 63.30\\
R1 & 100 & 4 & 48.00 & 54.50 & 68.50 & 72.00 & 82.83 & 60.75 & 65.17\\
R1 & 100 & 5 & 46.17 & 52.50 & 67.00 & 68.00 & 87.17 & 58.42 & 64.17\\
R1 & 100 & Mean & 44.63 & 53.33 & 65.23 & 69.07 & 79.80 & 58.07 & 62.41\\
\addlinespace
R2 & 100 & 1 & 42.67 & 53.17 & 58.17 & 69.00 & 79.33 & 55.75 & 60.47\\
R2 & 100 & 2 & 46.83 & 50.50 & 66.00 & 71.00 & 80.00 & 58.58 & 62.87\\
R2 & 100 & 3 & 46.17 & 53.83 & 66.50 & 71.17 & 82.17 & 59.42 & 63.97\\
R2 & 100 & 4 & 46.83 & 49.33 & 65.83 & 70.00 & 81.67 & 58.00 & 62.73\\
R2 & 100 & 5 & 46.83 & 53.67 & 65.33 & 72.83 & 86.67 & 59.67 & 65.07\\
R2 & 100 & Mean & 45.87 & 52.10 & 64.37 & 70.80 & 81.97 & 58.28 & 63.02\\
\addlinespace
\end{longtable}
\endgroup

\begingroup\small\setlength{\tabcolsep}{4pt}\setlength{\LTpre}{6pt}\setlength{\LTpost}{6pt}
\begin{longtable}{lllrrrrrr}
\caption{MobileNetV3-S, CIFAR-100: task accuracies during the final-task intervention at coefficients 10 and 100.}\label{tab:accuracy_epochs_mobilenetv3_small_cifar100}\\
\toprule
Branch & $\lambda$ & Epoch & T1 & T2 & T3 & T4 & Old & All\\\midrule
\endfirsthead
\multicolumn{9}{l}{\tablename\ \thetable\ (continued)}\\
\toprule
Branch & $\lambda$ & Epoch & T1 & T2 & T3 & T4 & Old & All\\\midrule
\endhead
\midrule\multicolumn{9}{r}{Continued on next page}\\\endfoot
\bottomrule\endlastfoot
Normal & 0 & 1 & 37.30 & 43.20 & 56.30 & 85.73 & 45.60 & 55.63\\
Normal & 0 & 2 & 36.33 & 40.87 & 57.30 & 84.77 & 44.83 & 54.82\\
Normal & 0 & 3 & 35.83 & 39.03 & 52.67 & 83.73 & 42.51 & 52.82\\
Normal & 0 & 4 & 35.07 & 38.97 & 51.57 & 84.67 & 41.87 & 52.57\\
Normal & 0 & 5 & 36.30 & 39.67 & 47.70 & 85.50 & 41.22 & 52.29\\
Normal & 0 & Mean & 36.17 & 40.35 & 53.11 & 84.88 & 43.21 & 53.63\\
\addlinespace
Scaffold & 10 & 1 & 38.13 & 41.20 & 65.53 & 75.63 & 48.29 & 55.12\\
Scaffold & 10 & 2 & 37.10 & 41.40 & 64.17 & 81.60 & 47.56 & 56.07\\
Scaffold & 10 & 3 & 40.03 & 43.33 & 63.73 & 83.73 & 49.03 & 57.71\\
Scaffold & 10 & 4 & 40.53 & 41.17 & 60.63 & 84.93 & 47.44 & 56.82\\
Scaffold & 10 & 5 & 38.77 & 42.00 & 59.53 & 85.50 & 46.77 & 56.45\\
Scaffold & 10 & Mean & 38.91 & 41.82 & 62.72 & 82.28 & 47.82 & 56.43\\
\addlinespace
R1 & 10 & 1 & 33.43 & 40.60 & 63.43 & 80.60 & 45.82 & 54.52\\
R1 & 10 & 2 & 37.23 & 39.93 & 62.73 & 85.40 & 46.63 & 56.32\\
R1 & 10 & 3 & 37.40 & 42.90 & 64.17 & 85.47 & 48.16 & 57.48\\
R1 & 10 & 4 & 37.80 & 43.53 & 63.17 & 85.50 & 48.17 & 57.50\\
R1 & 10 & 5 & 35.87 & 45.20 & 59.03 & 83.37 & 46.70 & 55.87\\
R1 & 10 & Mean & 36.35 & 42.43 & 62.51 & 84.07 & 47.10 & 56.34\\
\addlinespace
R2 & 10 & 1 & 33.83 & 40.70 & 63.73 & 80.80 & 46.09 & 54.77\\
R2 & 10 & 2 & 36.27 & 39.13 & 60.23 & 83.50 & 45.21 & 54.78\\
R2 & 10 & 3 & 36.70 & 40.63 & 61.17 & 82.83 & 46.17 & 55.33\\
R2 & 10 & 4 & 36.37 & 41.47 & 59.80 & 83.13 & 45.88 & 55.19\\
R2 & 10 & 5 & 37.90 & 43.43 & 60.63 & 85.23 & 47.32 & 56.80\\
R2 & 10 & Mean & 36.21 & 41.07 & 61.11 & 83.10 & 46.13 & 55.37\\
\addlinespace
Scaffold & 100 & 1 & 40.13 & 44.53 & 67.60 & 64.13 & 50.76 & 54.10\\
Scaffold & 100 & 2 & 42.47 & 44.43 & 65.63 & 72.30 & 50.84 & 56.21\\
Scaffold & 100 & 3 & 43.47 & 44.90 & 64.17 & 76.23 & 50.84 & 57.19\\
Scaffold & 100 & 4 & 42.87 & 45.57 & 65.03 & 78.67 & 51.16 & 58.03\\
Scaffold & 100 & 5 & 42.57 & 44.83 & 63.63 & 79.53 & 50.34 & 57.64\\
Scaffold & 100 & Mean & 42.30 & 44.85 & 65.21 & 74.17 & 50.79 & 56.64\\
\addlinespace
R1 & 100 & 1 & 35.20 & 41.90 & 68.07 & 67.23 & 48.39 & 53.10\\
R1 & 100 & 2 & 37.07 & 42.07 & 66.03 & 74.03 & 48.39 & 54.80\\
R1 & 100 & 3 & 39.60 & 45.60 & 64.90 & 76.93 & 50.03 & 56.76\\
R1 & 100 & 4 & 39.87 & 47.60 & 65.30 & 79.77 & 50.92 & 58.13\\
R1 & 100 & 5 & 41.30 & 46.50 & 63.90 & 79.43 & 50.57 & 57.78\\
R1 & 100 & Mean & 38.61 & 44.73 & 65.64 & 75.48 & 49.66 & 56.12\\
\addlinespace
R2 & 100 & 1 & 36.10 & 41.90 & 70.00 & 67.17 & 49.33 & 53.79\\
R2 & 100 & 2 & 38.03 & 42.87 & 67.00 & 74.63 & 49.30 & 55.63\\
R2 & 100 & 3 & 40.33 & 45.50 & 65.50 & 76.93 & 50.44 & 57.07\\
R2 & 100 & 4 & 39.20 & 43.67 & 64.90 & 76.67 & 49.26 & 56.11\\
R2 & 100 & 5 & 40.13 & 45.63 & 63.47 & 77.63 & 49.74 & 56.72\\
R2 & 100 & Mean & 38.76 & 43.91 & 66.17 & 74.61 & 49.62 & 55.86\\
\addlinespace
\end{longtable}
\endgroup

\begingroup\small\setlength{\tabcolsep}{4pt}\setlength{\LTpre}{6pt}\setlength{\LTpost}{6pt}
\begin{longtable}{lllrrrrrrr}
\caption{MLP-Mixer-B/16, CIFAR-10: task accuracies during the final-task intervention at coefficients 10 and 100.}\label{tab:accuracy_epochs_mixer_b16_cifar10}\\
\toprule
Branch & $\lambda$ & Epoch & T1 & T2 & T3 & T4 & T5 & Old & All\\\midrule
\endfirsthead
\multicolumn{10}{l}{\tablename\ \thetable\ (continued)}\\
\toprule
Branch & $\lambda$ & Epoch & T1 & T2 & T3 & T4 & T5 & Old & All\\\midrule
\endhead
\midrule\multicolumn{10}{r}{Continued on next page}\\\endfoot
\bottomrule\endlastfoot
Normal & 0 & 1 & 42.33 & 54.00 & 64.17 & 68.33 & 73.33 & 57.21 & 60.43\\
Normal & 0 & 2 & 42.17 & 54.17 & 57.67 & 68.33 & 81.33 & 55.58 & 60.73\\
Normal & 0 & 3 & 41.83 & 48.67 & 51.67 & 69.33 & 83.50 & 52.87 & 59.00\\
Normal & 0 & 4 & 44.00 & 56.00 & 57.00 & 66.00 & 87.00 & 55.75 & 62.00\\
Normal & 0 & 5 & 45.50 & 53.50 & 62.17 & 66.50 & 82.00 & 56.92 & 61.93\\
Normal & 0 & Mean & 43.17 & 53.27 & 58.53 & 67.70 & 81.43 & 55.67 & 60.82\\
\addlinespace
Scaffold & 10 & 1 & 50.00 & 54.00 & 51.00 & 32.50 & 65.33 & 46.88 & 50.57\\
Scaffold & 10 & 2 & 47.17 & 54.50 & 59.67 & 70.00 & 66.00 & 57.83 & 59.47\\
Scaffold & 10 & 3 & 50.67 & 53.67 & 58.33 & 71.33 & 70.33 & 58.50 & 60.87\\
Scaffold & 10 & 4 & 47.33 & 51.50 & 59.17 & 74.83 & 75.50 & 58.21 & 61.67\\
Scaffold & 10 & 5 & 48.67 & 53.83 & 60.33 & 74.50 & 76.50 & 59.33 & 62.77\\
Scaffold & 10 & Mean & 48.77 & 53.50 & 57.70 & 64.63 & 70.73 & 56.15 & 59.07\\
\addlinespace
R1 & 10 & 1 & 46.50 & 60.67 & 57.17 & 43.17 & 70.50 & 51.88 & 55.60\\
R1 & 10 & 2 & 47.83 & 56.00 & 58.67 & 67.33 & 74.50 & 57.46 & 60.87\\
R1 & 10 & 3 & 48.33 & 55.83 & 60.33 & 73.33 & 78.50 & 59.46 & 63.27\\
R1 & 10 & 4 & 46.00 & 54.50 & 60.17 & 71.83 & 82.50 & 58.13 & 63.00\\
R1 & 10 & 5 & 44.00 & 55.33 & 60.00 & 74.00 & 84.83 & 58.33 & 63.63\\
R1 & 10 & Mean & 46.53 & 56.47 & 59.27 & 65.93 & 78.17 & 57.05 & 61.27\\
\addlinespace
R2 & 10 & 1 & 51.33 & 59.67 & 59.33 & 43.33 & 67.33 & 53.42 & 56.20\\
R2 & 10 & 2 & 50.83 & 53.50 & 55.33 & 70.00 & 77.17 & 57.42 & 61.37\\
R2 & 10 & 3 & 50.33 & 56.67 & 58.00 & 73.17 & 81.83 & 59.54 & 64.00\\
R2 & 10 & 4 & 44.83 & 53.83 & 56.33 & 74.17 & 84.17 & 57.29 & 62.67\\
R2 & 10 & 5 & 44.00 & 51.17 & 55.67 & 72.50 & 87.33 & 55.83 & 62.13\\
R2 & 10 & Mean & 48.27 & 54.97 & 56.93 & 66.63 & 79.57 & 56.70 & 61.27\\
\addlinespace
Scaffold & 100 & 1 & 51.50 & 55.50 & 53.67 & 34.00 & 57.50 & 48.67 & 50.43\\
Scaffold & 100 & 2 & 44.17 & 48.83 & 54.83 & 68.17 & 61.67 & 54.00 & 55.53\\
Scaffold & 100 & 3 & 50.33 & 54.17 & 58.67 & 65.83 & 63.17 & 57.25 & 58.43\\
Scaffold & 100 & 4 & 48.50 & 53.00 & 55.83 & 73.17 & 69.00 & 57.62 & 59.90\\
Scaffold & 100 & 5 & 47.50 & 52.00 & 57.00 & 73.83 & 69.50 & 57.58 & 59.97\\
Scaffold & 100 & Mean & 48.40 & 52.70 & 56.00 & 63.00 & 64.17 & 55.02 & 56.85\\
\addlinespace
R1 & 100 & 1 & 40.83 & 49.83 & 42.50 & 21.67 & 71.83 & 38.71 & 45.33\\
R1 & 100 & 2 & 49.67 & 52.33 & 52.67 & 54.33 & 69.33 & 52.25 & 55.67\\
R1 & 100 & 3 & 48.83 & 51.50 & 54.17 & 57.00 & 71.67 & 52.87 & 56.63\\
R1 & 100 & 4 & 46.50 & 51.00 & 56.00 & 65.83 & 74.00 & 54.83 & 58.67\\
R1 & 100 & 5 & 46.17 & 50.83 & 56.33 & 69.00 & 74.67 & 55.58 & 59.40\\
R1 & 100 & Mean & 46.40 & 51.10 & 52.33 & 53.57 & 72.30 & 50.85 & 55.14\\
\addlinespace
R2 & 100 & 1 & 44.83 & 44.33 & 43.33 & 20.00 & 71.83 & 38.12 & 44.87\\
R2 & 100 & 2 & 47.83 & 50.50 & 53.00 & 54.50 & 70.00 & 51.46 & 55.17\\
R2 & 100 & 3 & 45.83 & 50.17 & 54.17 & 56.17 & 73.00 & 51.58 & 55.87\\
R2 & 100 & 4 & 47.00 & 51.67 & 55.50 & 65.00 & 73.50 & 54.79 & 58.53\\
R2 & 100 & 5 & 46.67 & 53.33 & 54.83 & 69.17 & 75.00 & 56.00 & 59.80\\
R2 & 100 & Mean & 46.43 & 50.00 & 52.17 & 52.97 & 72.67 & 50.39 & 54.85\\
\addlinespace
\end{longtable}
\endgroup

\begingroup\small\setlength{\tabcolsep}{4pt}\setlength{\LTpre}{6pt}\setlength{\LTpost}{6pt}
\begin{longtable}{lllrrrrrr}
\caption{MLP-Mixer-B/16, CIFAR-100: task accuracies during the final-task intervention at coefficients 10 and 100.}\label{tab:accuracy_epochs_mixer_b16_cifar100}\\
\toprule
Branch & $\lambda$ & Epoch & T1 & T2 & T3 & T4 & Old & All\\\midrule
\endfirsthead
\multicolumn{9}{l}{\tablename\ \thetable\ (continued)}\\
\toprule
Branch & $\lambda$ & Epoch & T1 & T2 & T3 & T4 & Old & All\\\midrule
\endhead
\midrule\multicolumn{9}{r}{Continued on next page}\\\endfoot
\bottomrule\endlastfoot
Normal & 0 & 1 & 33.67 & 38.20 & 58.40 & 85.07 & 43.42 & 53.83\\
Normal & 0 & 2 & 33.80 & 37.90 & 54.40 & 87.30 & 42.03 & 53.35\\
Normal & 0 & 3 & 30.40 & 34.10 & 52.20 & 86.57 & 38.90 & 50.82\\
Normal & 0 & 4 & 27.40 & 32.90 & 49.10 & 86.67 & 36.47 & 49.02\\
Normal & 0 & 5 & 31.13 & 34.37 & 49.77 & 87.50 & 38.42 & 50.69\\
Normal & 0 & Mean & 31.28 & 35.49 & 52.77 & 86.62 & 39.85 & 51.54\\
\addlinespace
Scaffold & 10 & 1 & 34.47 & 42.03 & 72.67 & 79.77 & 49.72 & 57.23\\
Scaffold & 10 & 2 & 33.80 & 41.30 & 70.57 & 86.43 & 48.56 & 58.02\\
Scaffold & 10 & 3 & 34.00 & 41.20 & 70.73 & 87.97 & 48.64 & 58.48\\
Scaffold & 10 & 4 & 34.47 & 41.30 & 69.27 & 90.20 & 48.34 & 58.81\\
Scaffold & 10 & 5 & 34.33 & 41.93 & 69.03 & 90.13 & 48.43 & 58.86\\
Scaffold & 10 & Mean & 34.21 & 41.55 & 70.45 & 86.90 & 48.74 & 58.28\\
\addlinespace
R1 & 10 & 1 & 30.77 & 37.27 & 63.57 & 87.20 & 43.87 & 54.70\\
R1 & 10 & 2 & 28.03 & 33.47 & 59.70 & 90.17 & 40.40 & 52.84\\
R1 & 10 & 3 & 27.70 & 32.83 & 55.97 & 90.33 & 38.83 & 51.71\\
R1 & 10 & 4 & 27.47 & 33.67 & 54.47 & 89.53 & 38.53 & 51.28\\
R1 & 10 & 5 & 29.07 & 33.63 & 55.63 & 89.33 & 39.44 & 51.92\\
R1 & 10 & Mean & 28.61 & 34.17 & 57.87 & 89.31 & 40.22 & 52.49\\
\addlinespace
R2 & 10 & 1 & 30.13 & 36.37 & 61.60 & 87.27 & 42.70 & 53.84\\
R2 & 10 & 2 & 28.20 & 34.83 & 59.73 & 89.97 & 40.92 & 53.18\\
R2 & 10 & 3 & 27.60 & 33.63 & 55.87 & 89.50 & 39.03 & 51.65\\
R2 & 10 & 4 & 27.53 & 36.33 & 57.37 & 89.93 & 40.41 & 52.79\\
R2 & 10 & 5 & 26.93 & 35.23 & 56.80 & 88.63 & 39.66 & 51.90\\
R2 & 10 & Mean & 28.08 & 35.28 & 58.27 & 89.06 & 40.54 & 52.67\\
\addlinespace
Scaffold & 100 & 1 & 35.53 & 41.57 & 72.70 & 73.80 & 49.93 & 55.90\\
Scaffold & 100 & 2 & 34.43 & 41.73 & 72.50 & 79.90 & 49.56 & 57.14\\
Scaffold & 100 & 3 & 34.37 & 41.70 & 71.77 & 81.77 & 49.28 & 57.40\\
Scaffold & 100 & 4 & 34.40 & 40.67 & 70.67 & 84.50 & 48.58 & 57.56\\
Scaffold & 100 & 5 & 34.17 & 40.73 & 70.70 & 85.07 & 48.53 & 57.67\\
Scaffold & 100 & Mean & 34.58 & 41.28 & 71.67 & 81.01 & 49.18 & 57.13\\
\addlinespace
R1 & 100 & 1 & 35.70 & 40.03 & 69.83 & 74.37 & 48.52 & 54.98\\
R1 & 100 & 2 & 33.20 & 37.93 & 69.13 & 83.53 & 46.76 & 55.95\\
R1 & 100 & 3 & 32.17 & 37.53 & 69.00 & 86.60 & 46.23 & 56.33\\
R1 & 100 & 4 & 30.63 & 36.07 & 66.83 & 89.03 & 44.51 & 55.64\\
R1 & 100 & 5 & 30.33 & 35.47 & 66.27 & 89.63 & 44.02 & 55.43\\
R1 & 100 & Mean & 32.41 & 37.41 & 68.21 & 84.63 & 46.01 & 55.67\\
\addlinespace
R2 & 100 & 1 & 35.63 & 40.30 & 69.07 & 74.83 & 48.33 & 54.96\\
R2 & 100 & 2 & 32.47 & 37.63 & 68.77 & 82.90 & 46.29 & 55.44\\
R2 & 100 & 3 & 31.40 & 38.33 & 69.30 & 85.83 & 46.34 & 56.22\\
R2 & 100 & 4 & 30.97 & 36.57 & 67.03 & 88.43 & 44.86 & 55.75\\
R2 & 100 & 5 & 31.40 & 37.17 & 67.23 & 89.80 & 45.27 & 56.40\\
R2 & 100 & Mean & 32.37 & 38.00 & 68.28 & 84.36 & 46.22 & 55.75\\
\addlinespace
\end{longtable}
\endgroup

\end{document}